\documentclass[11pt]{article}

\PassOptionsToPackage{table}{xcolor}
\usepackage[preprint]{acl}
\usepackage{times}
\usepackage{latexsym}
\usepackage[T1]{fontenc}
\usepackage[utf8]{inputenc}

\usepackage{microtype}
\usepackage{inconsolata}
\usepackage{graphicx}

\usepackage{amssymb}
\usepackage{amsmath}
\usepackage{booktabs}
\usepackage{multirow}
\usepackage{subcaption}

\title{iBERT: Interpretable Embeddings via Sense Decomposition}

\author{
 \textbf{Vishal Anand\textsuperscript{1}},
 \textbf{Milad Alshomary\textsuperscript{2}},
 \textbf{Kathleen McKeown\textsuperscript{2}}
\\
\\
 \textsuperscript{1}Microsoft, Washington, USA
 \\
 \textsuperscript{2}Columbia University, New York, USA
\\
 \small{
   \textbf{Correspondence:} \href{mailto:vishal.anand@microsoft.com}{vishal.anand@microsoft.com}
 }\\
 \small{\text{Website:} \href{https://iBERT.io}{https://iBERT.io}}
}

\usepackage{eso-pic}

\newcommand{\eaclstamp}{
  \AddToShipoutPictureFG*{
    \AtPageUpperLeft{
      \raisebox{-\height}[0pt][0pt]{
        \parbox{\paperwidth}{
          \vspace*{1.0em}{
          \hspace*{1.0em}\small
            Accepted to the Main Proceedings of EACL 2026
          }
        }
      }
    }
  }
}

\begin{document}
\eaclstamp
\maketitle

\begin{abstract}
\label{sec:abstract}
We present \textbf{iBERT} (interpretable-BERT), an encoder to produce inherently interpretable and controllable embeddings - designed to modularize and expose the discriminative cues present in language, such as semantic or stylistic structure. Each input token is represented as a sparse, non‑negative mixture over $k$ context‑independent \emph{sense vectors}, which can be pooled into sentence embeddings or used directly at the token level. This enables modular control over representation, before any decoding or downstream use.

To demonstrate our model's interpretability, we evaluate it on a suite of style‑focused tasks. On the STEL benchmark, it improves style representation effectiveness by $\sim$8 points over SBERT-style baselines, while maintaining competitive performance on authorship verification. Because each embedding is a structured composition of interpretable senses, we highlight how specific style attributes get assigned to specific sense vectors. While our experiments center on style, \mbox{iBERT} is not limited to stylistic modeling. Its structural modularity is designed to interpretably decompose whichever discriminative signals are present in the data — enabling generalization even when supervision blends semantic or stylistic factors.
\end{abstract}

\section{Introduction}
\label{sec:introduction}
\begin{figure}[t]
\centering
\includegraphics[width=\linewidth,trim={6.51cm 5.5cm 12.325cm 1.88cm},clip]{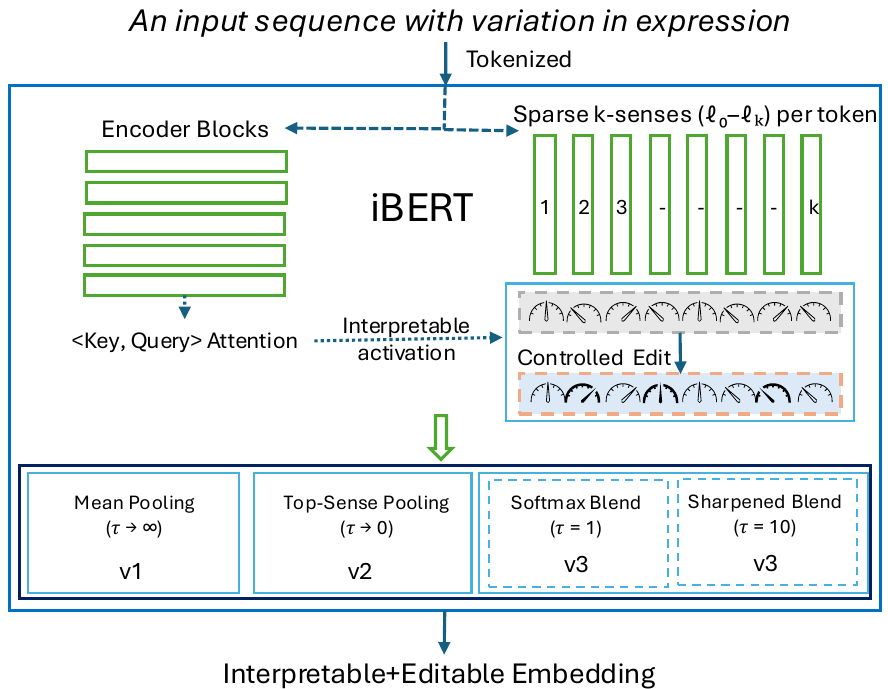}
\caption{iBERT encodes tokens via \textit{k} interpretable senses, producing editable and composable sense activations that are used either individually at the token level or pooled via configurable strategies into a global, interpretable embedding suitable for NLP pipelines.}
\label{fig:iBERT}
\vspace{-1.0em}
\end{figure}
Neural encoders increasingly serve as the backbone for tasks that rely on nuanced linguistic variation—ranging from authorship attribution and tone-controlled generation to stylistic retrieval and moderation. 
Yet, most popular encoders, such as SBERT \cite{reimers-gurevych-2019-sentence} or SimCSE \cite{gao-etal-2021-simcse}, produce dense vector representations that offer no clear control over how style and meaning are encoded. This limits their reliability in domains where representational transparency and stylistic control are essential. These challenges are especially acute in stylistic tasks, where disentangling content from style is not optional but required. Without structured representations, it becomes difficult to attribute stylistic effects, audit model behavior, or intervene in generation pipelines. Prior approaches often rely on proxy supervision and post-hoc explanations, which yield entangled embeddings and incomplete attribution 
\cite{danilevsky-etal-2020-survey, john-etal-2019-disentangled, elazar-etal-2021-amnesic}.

In this work, we ask: 
\textit{Can we design an encoder where the semantic and stylistic representations are made explicit and controllable within the embedding space—by design, not post-hoc?}
To this end, we develop \textbf{iBERT}, an encoder architecture with representations that are interpretable and controllable by design. Each token is expressed as a sparse, non-negative mixture over $k$ context-independent \textit{sense vectors}, which can be pooled into sentence embeddings or used directly at the token level. This design enables modular control over specific dimensions of meaning and style that are learned during the training phase, enabling analysis, attribution, and targeted edits in embedding space. Our architecture builds on the Backpack formulation of \citet{hewitt-etal-2023-backpack}, which modeled autoregressive decoding using sparse token-level senses. We adapt this formulation to an encoder-only setting with global pooling, allowing for sentence-level compositionality, bidirectional input, and plug-and-play use across classification and retrieval pipelines.

We train iBERT with a masked language modeling objective on a web corpus to produce sparse, interpretable sense embeddings.
We evaluate iBERT on style analysis tasks—benchmarks that test a model’s ability to isolate, attribute, and manipulate stylistic variation without entangling semantic content. In particular, by training iBERT on the style analysis task \cite{patel-etal-2025-styledistance, wegmann-etal-2022-author}, we find that it improves style representation accuracy (STEL) by +8 points over SBERT-style baselines while remaining competitive on Style-or-Content (SoC) and PAN authorship verification. Because each embedding is a transparent mixture of sense vectors, we can directly identify stylistic axes (e.g., emoji use, sarcasm, lexical choices) and apply targeted edits in the embedding space.

Though evaluated on stylistic tasks, iBERT is not a style model. It is a general-purpose encoder with decomposable representations—suitable for domains where interpretability and control are prerequisites rather than afterthoughts.

\paragraph{Contributions.}
\begin{itemize}
    \vspace{-0.2em}
    \item We present \textbf{iBERT}, an encoder with controllable decomposable representations, enabling interpretable token and sentence embeddings.

    \vspace{-0.3em}
    \item We show strong performance across three benchmarks: STEL, SoC, and PAN - achieving up to \textbf{+8\% STEL} gains over baselines.

    \vspace{-0.3em}
    \item We demonstrate novel interpretability capabilities: probing sense vectors for stylistic traits, ablating senses to identify attribution, and editing embeddings for controlled style transfer.
\end{itemize}

\section{Background and Related Work}
\label{sec:related-work}

\paragraph{Sentence encoders and interpretability.}
Transformer-based encoders such as SBERT \cite{reimers-gurevych-2019-sentence} and SimCSE \cite{gao-etal-2021-simcse} perform well on semantic retrieval and sentence similarity tasks, but their dense embeddings offer little visibility into what linguistic attributes are being captured.  
Efforts to explain these representations have focused on post-hoc methods such as probing-based explanations \cite{wallace-etal-2019-allennlp}, saliency and attribution techniques \cite{jacovi-goldberg-2020-towards}, and influence functions.  
As noted by \citet{jacovi-goldberg-2020-towards}, such methods are often applied to representations that were not designed for interpretability, and can yield incomplete or unfaithful explanations.  
The survey by \citet{danilevsky-etal-2020-survey} further emphasizes the need for \emph{architecturally grounded} interpretability—where structure and transparency are embedded into the model itself.

\paragraph{Stylistic representation learning.}
Most prior work on stylistic representation relies on \emph{proxy supervision}, typically using authorship identity as a surrogate for stylistic consistency.  
For example, \citet{wegmann-etal-2022-author} construct contrastive triplets using same-author, different-topic texts to encourage content-invariant representations, and evaluate model performance using metrics such as STEL and Style-or-Content (SoC).
While effective, these approaches still produce dense, entangled embeddings with no inherent interpretability or mechanism for stylistic control.

\citet{patel-etal-2025-styledistance} introduce a new paradigm: supervising \emph{style directly}, independent of topic, by generating contrastive triplets with controlled lexical and syntactic variation.  
Their StyleDistance model, along with the SynthSTEL benchmark, enables direct evaluation along 40 distinct content-controlled style-axes.
We adopt this direct-style supervision setup, and pair it with our new architectural formulation: an encoder that produces decomposable, interpretable embeddings by design.

\paragraph{Multi-sense representations and Backpack.}
Backpack-GPT \cite{hewitt-etal-2023-backpack} introduced sparse token-level representations via mixtures over static senses, enabling interpretable and locally steerable generation in autoregressive models.  
While their formulation supports tokenwise control, it operates within a decoder-centric setup and focuses on in-loop manipulation during generation.

iBERT repositions sparse sense mixtures as a representation-level mechanism, producing globally composable embeddings that enable interpretability and control across tasks.
This allows for interpretable, task-agnostic control at the representation level—prior to any decoding—making \mbox{iBERT} compatible with retrieval, classification, and generation pipelines alike.  
Rather than serving as a generation-side control mechanism, iBERT reframes sense-based modeling as a general-purpose representational interface for controllable NLP.

\section{Method}
\label{sec:method}
We present \textbf{iBERT}, a multi-sense encoder that produces interpretable embeddings by design. iBERT can operate as a token-level encoder (pretrained as MLM) or as a sentence encoder (v1–v3) trained with contrastive supervision. Each token is encoded as a sparse mixture over static \emph{sense vectors}, which are then pooled to form globally composable sentence embeddings. This design enables explicit analysis, control, and manipulation of embeddings. In the paper, the usage of term BERT in a sentential context implies Sentence-BERT \cite{reimers-gurevych-2019-sentence}.

\vspace{0.5em}
\subsection{iBERT Architecture}

\autoref{fig:iBERT} shows a high-level overview of our architecture; a full schematic is provided in the Appendix (\autoref{fig:iBERT-detailed}). Each token $x_i$ is first mapped to a standard embedding $e_i = E\,x_i \in \mathbb{R}^d$. A feed-forward layer projects this into $k$ \textbf{senses}:
\[
C(x_i) = [c(x_i)_1, \dots, c(x_i)_k] \in \mathbb{R}^{k \times d}
\]

In parallel, a transformer encoder produces contextual states $H = [h_1, \dots, h_n] \in \mathbb{R}^{n \times d}$. These are projected into sense-specific queries and keys: $Q^{(\ell)}, K^{(\ell)} \in \mathbb{R}^{(d/k) \times d}$, allowing us to compute mixture weights $\alpha_{\ell,i,j}$—the contribution of sense $\ell$ from token $x_j$ to the contextual embedding of token $x_i$.

\vspace{-2.0em}
\[
o_i = \sum_{j=1}^{n} \sum_{\ell=1}^{k} \alpha_{\ell,i,j} \cdot c(x_j)_\ell, \quad o_i \in \mathbb{R}^d.
\]

The resulting $o_i$ is a convex mixture of context-independent sense vectors. These representations can be inspected or edited directly. We pretrain this architecture as iBERT-MLM with $k$=8 senses using masked language modeling, yielding a 171M-parameter encoder, comparable in scale to BERT-base. At k=1, the sense construction block reduces to an embedding matrix, since each token is now mapped to a traditional one dimensional vector.

\vspace{0.5em}
\subsection{iBERT Sentence Embeddings}
\label{subsection:iBERT-sentence-embeddings}
As a general idea, we take the two-dimensional sense vectors per token and construct a weighted structure to condense the senses together. We apply structured pooling over each token\textsubscript{i} representation $o_i$, resulting in three variants: v1 (uniform averaging), v2 (top-sense selection), and v3 (soft blending senses wired-into the network by a sense composition function).

\paragraph{iBERT-v1: Mean pooling.}
The sentence embedding is computed via uniform averaging:
\vspace{-0.5em}
\[
s = \frac{1}{n} \sum_{i=1}^{n} o_i.
\]

This produces a decomposable embedding, where each dimension reflects contributions from token-level sense activations—supporting interpretation and attribution (Figure~\ref{fig:iBERT-v1}).

\paragraph{iBERT-v2: Top-sense pooling.}
For an input sequence, we compute the total activation per-sense:\vspace{-0.3em}
\[
S_\ell = \sum_{i=1}^{n} \left\| o_i^{(\ell)} \right\|_2, \text{where } o_i^{(\ell)} = \sum_j \alpha_{\ell,i,j} \cdot c(x_j)_\ell
\]
We identify the dominant sense (\textit{per sequence}) as $\ell^* = \arg\max_\ell S_\ell$, and retain only that sense across all tokens to compute the sequence embedding:
\[
s = \frac{1}{n} \sum_{i=1}^{n} o_i^{(\ell^*)}.
\]
This encourages stricter top-1 alignment between input sequences and individual senses. The process is repeated independently for each input sequence (Figure~\ref{fig:iBERT-v2}).

\paragraph{iBERT-v3: Softmax-weighted pooling.}
We define a general family of models to interpolate behaviors between v1 and v2 using softmax over sense norms, controlled by a sense composition variable $\tau$. Once $\tau$ is set, the v3 trained model is fixed with the pooling structure and is not replaced afterwards.
\vspace{-0.5em}
\[
S_\ell = \sum_{i=1}^{n} \left\| o_i^{(\ell)} \right\|_2, \quad
\pi_\ell(\tau) = \mathrm{softmax}(S_\ell / \tau),
\]
\vspace{-0.8em}
\[
s_\tau = \frac{1}{n} \sum_{i=1}^{n} \sum_{\ell=1}^{k} \pi_\ell(\tau) \cdot o_i^{(\ell)}.
\]
This general formulation defines v3, where:
\begin{itemize}
  \vspace{-0.4em}
  \item $\tau \rightarrow \infty$: recovers v1 (uniform averaging)
  \vspace{-0.1em}
  \item $\tau \rightarrow 0$: recovers v2 (hard top-sense selection)
  \item $0 < \tau < \infty$: soft blending across senses
\end{itemize}

This generalized framework allows us to study how the sharpness of sense composition affects interpretability and performance (\S\ref{sec:exp}, Appendix~\ref{app:tau}).

\subsection{Training Pipeline}
\label{sec:training}

We train iBERT in two stages. Unlike prior work \cite{patel-etal-2025-styledistance, wegmann-etal-2022-author} which fine-tuned off-the-shelf encoders, our decomposable design requires MLM pretraining to instantiate meaningful sense vectors.

\subsubsection{Stage 0: MLM Pretraining.}
\label{sec:mlm-pretraining}
We pretrain iBERT-MLM from scratch on 5\% of \textsc{FineWeb} \cite{10.5555/3737916.3738886}, a 15T-token web-scale corpus. This 750B-token slice offers broad linguistic and stylistic diversity while remaining computationally efficient. We use standard masked language modeling to learn token-level sense activations, producing a 171M-parameter encoder. For comparison, we also train a standard BERT model using identical settings and data.

\subsubsection{Stage 1: Style Representaion Learning}
To train iBERT on the task of style analysis, we use contrastive triplets $\langle s, s^+, s^- \rangle$ targeting stylistic variation and create 
iBERT-v1-v3. Training data is drawn from \textit{StyleSynth} by \citet{patel-etal-2025-styledistance}: content-controlled, synthetic style triplets; and from author-labeled triplets with topic control by \citet{wegmann-etal-2022-author}.

We apply the InfoNCE loss to encourage closeness between $s$ and $s^+$, while pushing $s$ away from $s^-$. Pooling weights $\pi_\ell(\tau)$ are shared between the anchor and positive, encouraging consistent sense-axis usage for stylistically similar inputs.

To ensure architectural parity, we also train \emph{BERT} and then \emph{SBERT} baselines using the two-stage setup: a BERT-base model pretrained on the same 5\% FineWeb slice, followed by contrastive fine-tuning on anchored triplets. This controls for pretraining variance and isolates the contribution of our multi-sense architecture.

\paragraph{Model size.}
iBERT-v1–v3 has 171.4M parameters and SBERT has 149.7M. Inference is approximately 2\% slower owing to sense composition. Tradeoffs are discussed in Section \ref{sec:limitations}.

\section{Experimental Setup}
\label{sec:exp}

\subsection{Pretraining and Fine-tuning}
We reuse iBERT-MLM and BERT models from Stage 0 (\S\ref{sec:mlm-pretraining}), trained from scratch on 5\% of \textsc{FineWeb}. Sentence-level variants—iBERT-v1–v3 are fine-tuned on contrastive triplets to learn style representation using either:

\begin{itemize}
  \item \textbf{StyleDistance - SD} \cite{patel-etal-2025-styledistance}: 50k synthetic triplets (StyleSynth
  ) with controlled lexical and syntactic variation. Median token-size is 19 (Appendix \autoref{tab:token_stats}).
  \vspace{-0.4em}
  \item \textbf{Wegmann - WG} \cite{wegmann-etal-2022-author}: 40k author-anchored Reddit triplets with similar topics. Median token-length is 24.
\end{itemize}
\vspace{-0.3em}
Training follows the loss function of each dataset: for StyleSynth we use the InfoNCE objective as in \citet{patel-etal-2025-styledistance}, while for Wegmann triplets we apply the margin-based contrastive loss used in \citet{wegmann-etal-2022-author}. In both cases, pooling weights $\pi_\ell(\tau)$ are shared between the anchor and positive to enforce consistent sense usage. Additional details are in Appendix~\ref{app:impl}.

\paragraph{Baselines.}
To isolate the effect of our multi-sense architecture, we compare against contrastively trained versions of the same BERT encoder from Stage 0.  
We fine-tune this model to create:
\begin{itemize}
  \vspace{-0.4em}
  \item BERT\textsubscript{SD}: trained on StyleSynth triplets with InfoNCE loss (as in \citet{patel-etal-2025-styledistance});
  \vspace{-0.4em}
  \item BERT\textsubscript{WG}: trained on Wegmann triplets with the margin-based triplet loss (as in \citet{wegmann-etal-2022-author});
  \vspace{-0.4em}
  \item BERT\textsubscript{WG+SD}: trained on both datasets with InfoNCE loss (reflecting the strongest setup reported by \citet{patel-etal-2025-styledistance}), which is listed as the prior state-of-the-art performing model called StyleDistance.
\end{itemize}
BERT\textsubscript{WG} corresponds to the SBERT-style model of \citet{wegmann-etal-2022-author}, while BERT\textsubscript{WG+SD} mirrors the best-performing StyleDistance setup of \citet{patel-etal-2025-styledistance}.
All settings share the same backbone, initialization, and optimizer for a controlled comparison.

\paragraph{Backbone Consistency.}
Prior work used DistilRoBERTa-based encoders (e.g., \citet{patel-etal-2025-styledistance}, \citet{wegmann-etal-2022-author}). Since iBERT is trained from scratch, we also train a new BERT-base encoder using the same setup as iBERT-MLM.
This serves as the backbone for BERT\textsubscript{SD / WG / WG+SD}, replacing DistilRoBERTa in their respective setups and enabling a fair performance comparison. Full training protocol is detailed in Appendix~\ref{app:impl}.

\subsection{Evaluation Benchmarks}

We evaluate all models across four direct style tasks (STEL-40\textsubscript{\textsc{SD}}, SoC-40\textsubscript{\textsc{SD}}, STEL-5\textsubscript{\textsc{WG}}, SoC-5\textsubscript{\textsc{WG}}) and one proxy-style task (Authorship Verification - PAN), using both 128- and 512-token variants of the backbone BERT models.

\paragraph{Direct-style evaluation.}
\begin{itemize}
  \item \textbf{STEL}: multi-class classification over 40 style labels (StyleSynth) and 5 style classes (Wegmann), reported as weighted accuracy.
  \item \textbf{SoC}: average binary classification accuracy between positive/negative style polarities per class (e.g., formal vs. informal).
\end{itemize}

\paragraph{Proxy-style evaluation.}
\textbf{PAN} authorship verification: binary classification of same vs. different author pairs (PAN 2011/13/14/15), reporting AUC. Treated as a proxy for style consistency, though not grounded in specific stylistic dimensions.

\section{Results}
\label{sec:results}

\subsection{STEL Performance}
\label{sec:res:synthstel}
\begin{table*}[t]
\centering
\setlength{\tabcolsep}{5pt}
\resizebox{0.78\textwidth}{!}{
\begin{tabular}{@{}cl|ccc|ccc@{}}
\toprule
\multirow{2.5}{*}{\small\textsc{Data}} 
& \multirow{2.5}{*}{\textsc{Models}} & \multicolumn{3}{c}{128 tokens} & \multicolumn{3}{c}{512 tokens} \\
\cmidrule(lr){3-5}\cmidrule(lr){6-8}
& & STEL$\uparrow$ & SoC$\uparrow$ & PAN$\uparrow$ & STEL$\uparrow$ & SoC$\uparrow$ & PAN$\uparrow$ \\
\midrule
\midrule
\multirow{5}{*}{\cellcolor{white}\rotatebox{90}{\small \textsc{SD-only}}}
& BERT & 31.6 $\pm$ {\small 0.78} & 91.5 $\pm$ {\small 0.56} & \textit{60.57} & 30.6 $\pm$ {\small 0.86} & \textit{91.7} $\pm$ {\small 0.33} & \underline{\textbf{63.52}} \\
& \cellcolor{gray!12}iBERT-v1 & \cellcolor{gray!12}\textit{37.5} $\pm$ {\small 0.41} & \cellcolor{gray!12}\textit{92.3} $\pm$ {\small 0.33} & \cellcolor{gray!12}\textbf{60.88} & \cellcolor{gray!12}\textbf{\underline{38.9}} $\pm$ {\small 0.81} & \cellcolor{gray!12}\textit{92.0} $\pm$ {\small 0.29} & \cellcolor{gray!12}59.49 \\
& \cellcolor{gray!12}iBERT-v2 & \cellcolor{gray!12}28.5 $\pm$ {\small 1.13} & \cellcolor{gray!12}88.7 $\pm$ {\small 0.68} & \cellcolor{gray!12}58.95 & \cellcolor{gray!12}31.4 $\pm$ {\small 1.46} & \cellcolor{gray!12}91.1 $\pm$ {\small 0.68} & \cellcolor{gray!12}59.41 \\
& \cellcolor{gray!12}iBERT-v3-1 & \cellcolor{gray!12}\textbf{38.3} $\pm$ {\small 0.62} & \cellcolor{gray!12}\textbf{\underline{92.8}} $\pm$ {\small 0.48} & \cellcolor{gray!12}58.15 & \cellcolor{gray!12}\textit{38.2} $\pm$ {\small 0.75} & \cellcolor{gray!12}\textit{92.0} $\pm$ {\small 0.32} & \cellcolor{gray!12}59.39 \\
& \cellcolor{gray!12}iBERT-v3-10 & \cellcolor{gray!12}\textit{38.1} $\pm$ {\small 0.44} & \cellcolor{gray!12}\textit{92.5} $\pm$ {\small 0.30} & \cellcolor{gray!12}58.29 & \cellcolor{gray!12}36.0 $\pm$ {\small 0.21} & \cellcolor{gray!12}\textbf{\underline{92.0}} $\pm$ {\small 0.30} & \cellcolor{gray!12}59.41 \\
\addlinespace[2pt]
\hline
\addlinespace[2pt]
\multirow{5}{*}{\cellcolor{white}\rotatebox{90}{\small \textsc{WG+SD}}}
& BERT & 30.6 $\pm$ {\small 0.93} & 88.9 $\pm$ {\small 0.53} & 58.96 & 30.4 $\pm$ {\small 0.94} & \textit{89.6} $\pm$ {\small 0.38} & \textbf{62.76} \\
& \cellcolor{gray!12}iBERT-v1 & \cellcolor{gray!12}\textbf{\underline{38.6}} $\pm$ {\small 1.57} & \cellcolor{gray!12}\textit{90.0} $\pm$ {\small 0.22} & \cellcolor{gray!12}58.96 & \cellcolor{gray!12}\textit{35.7} $\pm$ {\small 1.76} & \cellcolor{gray!12}\textit{89.5} $\pm$ {\small 0.70} & \cellcolor{gray!12}\textit{61.53} \\
& \cellcolor{gray!12}iBERT-v2 & \cellcolor{gray!12}33.4 $\pm$ {\small 0.54} & \cellcolor{gray!12}89.1 $\pm$ {\small 0.50} & \cellcolor{gray!12}58.47 & \cellcolor{gray!12}32.3 $\pm$ {\small 1.13} & \cellcolor{gray!12}89.0 $\pm$ {\small 0.54} & \cellcolor{gray!12}59.54 \\
& \cellcolor{gray!12}iBERT-v3-1 & \cellcolor{gray!12}\textit{38.0} $\pm$ {\small 0.85} & \cellcolor{gray!12}\textbf{90.2} $\pm$ {\small 0.30} & \cellcolor{gray!12}58.56 & \cellcolor{gray!12}\textbf{36.4} $\pm$ {\small 0.71} & \cellcolor{gray!12}\textbf{90.1} $\pm$ {\small 0.58} & \cellcolor{gray!12}\textit{61.43} \\
& \cellcolor{gray!12}iBERT-v3-10 & \cellcolor{gray!12}\textit{38.0} $\pm$ {\small 0.88} & \cellcolor{gray!12}\textit{89.6} $\pm$ {\small 0.65} & \cellcolor{gray!12}\textbf{\underline{60.96}} & \cellcolor{gray!12}34.8 $\pm$ {\small 0.93} & \cellcolor{gray!12}\textit{89.8} $\pm$ {\small 0.32} & \cellcolor{gray!12}60.34 \\
\addlinespace[2pt]
\hline
\addlinespace[2pt]
\multirow{5}{*}{\cellcolor{white}\rotatebox{90}{\small \textsc{WG-only}}}
& BERT & 27.1 $\pm$ {\small 0.70} & \textbf{27.5} $\pm$ {\small 1.18} & 58.20 & \textbf{27.3} $\pm$ {\small 0.97} & \textbf{28.9} $\pm$ {\small 1.53} & \textbf{61.92} \\
& \cellcolor{gray!12}iBERT-v1 & \cellcolor{gray!12}\textit{28.5} $\pm$ {\small 0.86} & \cellcolor{gray!12}8.0 $\pm$ {\small 0.62} & \cellcolor{gray!12}\textit{59.29} & \cellcolor{gray!12}\textit{27.2} $\pm$ {\small 0.74} & \cellcolor{gray!12}8.0 $\pm$ {\small 0.73} & \cellcolor{gray!12}60.74 \\
& \cellcolor{gray!12}iBERT-v2 & \cellcolor{gray!12}25.9 $\pm$ {\small 0.76} & \cellcolor{gray!12}6.7 $\pm$ {\small 0.65} & \cellcolor{gray!12}57.77 & \cellcolor{gray!12}\textit{26.8} $\pm$ {\small 0.56} & \cellcolor{gray!12}6.3 $\pm$ {\small 0.75} & \cellcolor{gray!12}58.63 \\
& \cellcolor{gray!12}iBERT-v3-1 & \cellcolor{gray!12}\textit{28.1} $\pm$ {\small 0.59} & \cellcolor{gray!12}7.4 $\pm$ {\small 0.42} & \cellcolor{gray!12}\textit{59.47} & \cellcolor{gray!12}\textit{27.1} $\pm$ {\small 0.73} & \cellcolor{gray!12}7.4 $\pm$ {\small 0.50} & \cellcolor{gray!12}60.87 \\
& \cellcolor{gray!12}iBERT-v3-10 & \cellcolor{gray!12}\textbf{28.5} $\pm$ {\small 0.58} & \cellcolor{gray!12}7.3 $\pm$ {\small 0.41} & \cellcolor{gray!12}\textbf{59.87} & \cellcolor{gray!12}26.5 $\pm$ {\small 0.96} & \cellcolor{gray!12}7.4 $\pm$ {\small 0.68} & \cellcolor{gray!12}\textit{61.07} \\
\bottomrule
\end{tabular}
}
\caption{Direct-style: STEL and SoC reported on the \textbf{45 style groups} (40 of StyleSynth + 5 of Wegmann STEL). Proxy-style: PAN reports average AUC\% on PAN11/13/14/15 authorship verification tasks. The models are trained on corresponding \textsc{Data} triplets for \textbf{5 runs}. Median token-lengths: 19 for SD, and 24 for WG (Appendix \autoref{tab:token_stats}). BERT$_{\text{WG+SD}}$, named StyleDistance, is the prior state-of-the-art setup.}
\label{tab:synthstel-45}
\vspace{-0.8em}
\end{table*}

\autoref{tab:synthstel-45} reports style representation effectiveness (STEL) across training regimes \cite{wegmann-nguyen-2021-capture}. Across the board, iBERT models outperform their BERT counterparts, underscoring the utility of structured, interpretable pooling for capturing stylistic signals. Under \textsc{SD-only} training, iBERT-v3-1 achieves the highest STEL scores, yielding 38.3\% (128 tokens) and 38.2\% (512 tokens)—a substantial gain of +6.7 and +7.6 points over BERT. Close contenders include iBERT-v1 and iBERT-v3-10, both of which exhibit similarly strong performance, suggesting that soft, sense-weighted pooling is especially effective when guided by direct-style supervision.

Interestingly, iBERT-v2 consistently lags behind v1 and v3 across all settings. This is expected: its hard top-sense selection mechanism introduces sharp sparsity that likely harms generalization by collapsing compositional diversity. In contrast, iBERT-v3-10 is not only among the top performers but also the most stable across settings—offering a smoother inductive bias that bridges the extremes of v1 (uniform blend) and v2 (hard top-1 sense).

Even under joint supervision from \textsc{WG+SD}, where noisy author-level triplets may dilute sense-specific stylistic signal, iBERT variants maintain their edge. iBERT-v1 secures the best STEL at 128 tokens (38.6\%), while iBERT-v3-1 leads at 512 tokens (36.4\%), outperforming BERT by approximately 6 points in both cases.

Finally, in the \textsc{WG-only} setting, where labels reflect author identity rather than explicit styles, all models see degraded performance. Nevertheless, iBERT-v3 and iBERT-v1 is at par with BERT, signaling that even without style supervision, iBERT retains competitive structure-aware generalizability. These results affirm that iBERT's compositional inductive bias enables robustness without compromising direct-style effectiveness.

\subsection{SoC Performance}
For the SoC metric—evaluating separability of stylistic polarity—iBERT-v3-1 achieves the highest overall scores: 92.8\% (128 tokens) and 92.0\% (512 tokens) under \textsc{SD-only}, outperforming all baselines. These results show that iBERT can preserve or even enhance polarity sensitivity while maintaining interpretability.

In contrast, iBERT-v2 underperforms across settings (e.g., 88.7\% under \textsc{SD-only}), often lagging behind both v1 and BERT. We attribute this to its top-1 sense pooling: by forcing each sequence to align with a single dominant sense, iBERT-v2 severely restricts the number of senses that can specialize for the same data-points observed. This architectural bottleneck reduces expressive capacity, especially for fine-grained polarity distinctions that require distributed compositionality.

Under \textsc{WG-only} training—where supervision is provided weakly via author labels rather than direct stylistic polarity—all models see a dramatic drop in SoC performance. BERT, for example, falls from 91.5\% (\textsc{SD-only}) to just 27.5\%, while iBERT variants drop even further (e.g., 8.0\% for v1). Despite this collapse, BERT appears to outperform iBERT in this setting. We hypothesize this is due to \textit{content leakage}: BERT implicitly entangles semantic and stylistic attributes, using topic or lexical cues as proxies for style. While this can yield fragile SoC signal in weakly supervised settings, it sacrifices interpretability and modularity. In contrast, iBERT's explicit disentanglement fails to recover polarity distinctions without true stylistic supervision—a tradeoff aligned with its core design goals.

Taken together, these results affirm that iBERT excels when trained with style-specific signals, achieving strong polarity separability without compromising compositional control or interpretability. Further analysis of the pooling sharpness mechanism (\S \ref{subsection:iBERT-sentence-embeddings}) translating to performance and stability is discussed in Appendix \ref{app:tau}.

\subsection{Authorship Verification (PAN)}
\label{sec:pan}
To assess generalizability of learning style to solve the authorship attribution task, we use the PAN11–15 benchmarks, a suite of proxy-style tasks requiring models to distinguish between same-author and different-author text pairs. Unlike STEL or SoC, PAN tasks blend stylistic and semantic variation, providing a rich probe of compositional generalization. The results are in Tables \ref{tab:synthstel-45} and \ref{tab:pan_sd_wg_128_512}.

WG training dataset is a mixed dataset, containing author identity supervision with intertwined semantic and stylistic signals. Unlike BERT, which entangles all available cues, iBERT is architecturally designed to tease apart these differentials in an interpretable fashion. This enables it to maintain competitive performance even when the supervision is noisy or not explicitly style-specific—while still excelling on clean, style-focused tasks like STEL and SoC.

At 128 tokens, all iBERT variants match or slightly outperform BERT, with iBERT-v3-10 achieving the highest AUC (59.87\%) for \textsc{WG}. This is notable not because iBERT ignores semantic information, but because it selectively organizes it in disentangled form—generalizing well even when stylistic cues are weakly defined.

At 512 tokens, BERT slightly outperforms \mbox{iBERT} (61.92\% vs. 61.07\% for v3), with the largest margins on PAN13 and PAN15, that are characterized by semantic drift and cross-topic variation. This suggests that BERT benefits more from extended context, due to its architecture's tight coupling of semantic and stylistic features. In contrast, iBERT's modular structure—designed to isolate stylistic axes—does not directly exploit semantic continuity. Still, the gap stays slim, highlighting the semantic signals acquired in MLM stage remain accessible even in disentangled representations.

Interestingly, even iBERT-v2, which lagged on STEL and SoC, performs competitively on PAN. This underscores that semantic signals are preserved, even when stylistic specialization is weak. Looking across datasets, iBERT performs best on PAN11 and PAN14, where stylistic consistency is high and maintains near-parity on PAN13 and PAN15, where semantic variation dominates.

While \textsc{WG} provides author identity supervision, its examples entangle stylistic and semantic cues: offering iBERT no clean signals to isolate. In contrast, \textsc{SD} supplies disentangled stylistic supervision but lacks authorial labels entirely. This creates a supervision-task mismatch for iBERT, which is architecturally primed to extract interpretable signals. Yet it adapts: from WG, it organizes conflated authorial traits (e.g., tone, topic, lexical patterns) into structured axes; from SD, it learns clean stylistic subspaces that can approximate the mixed style-content clusters found in PAN’s test sets. That iBERT performs strongly across PAN benchmarks—despite this misalignment—demonstrates its ability to decompose and generalize whichever discriminative signals are available, even when style and semantics are blended.

\subsection{Interpretability Analysis of iBERT}
\begin{table}[t]
\centering\small
\resizebox{0.48\textwidth}{!}{
\begin{tabular}{c l c}
\toprule
\textsc{\textbf{Sense}} & \textsc{\textbf{Top-Aligned Style Axes}} & \textsc{\textbf{Emergent Theme}} \\
\midrule

\multirow{3}{*}{$\ell=0$}
  & \textit{With Emojis / No Emojis} & \multirow{3}{*}{\shortstack[c]{Surface-level\\markers}} \\
  & \textit{Frequent / Infrequent Conjunctions} & \\
  & \textit{Frequent / Infrequent Personal Pronouns} & \\[0.5ex]

\midrule
\multirow{4}{*}{$\ell=1$}
  & \textit{All Upper Case / Proper Capitalization} & \multirow{4}{*}{\shortstack[c]{Orthographic and\\visual style}} \\
  & \textit{Text Emojis / No Emojis} & \\
  & \textit{Long / Short Average Word Length} & \\
  & \textit{With / Without Number Substitution} & \\[0.5ex]

\midrule
\multirow{4}{*}{$\ell=2$}
  & \textit{Humorous / Non-Humorous} & \multirow{4}{*}{\shortstack[c]{Affect and\\expressive tone}} \\
  & \textit{Sarcastic / Non-Sarcastic} & \\
  & \textit{Metaphoric / Literal} & \\
  & \textit{Offensive / Non-Offensive} & \\[0.5ex]

\midrule
\multirow{2}{*}{$\ell=3$}
  & \textit{All Lower Case / Proper Capitalization} & \multirow{2}{*}{\shortstack[c]{Textual correctness\\and noise}} \\
  & \textit{With Misspellings / Normal Sentence} & \\[0.5ex]

\midrule
\multirow{2}{*}{$\ell=4$}
  & \textit{More / Less Frequent Function Words} & \multirow{2}{*}{Functional grammar} \\
  & \textit{With / Without Nominalizations} & \\[0.5ex]

\midrule
\multirow{2}{*}{$\ell=5$}
  & \textit{Active / Passive} & \multirow{2}{*}{\shortstack[c]{Syntactic voice\\and register}} \\
  & \textit{Contracted / Non-Contracted} & \\[0.5ex]

\midrule
\multirow{2}{*}{$\ell=6$}
  & \textit{Frequent / Infrequent Pronouns} & \multirow{2}{*}{\shortstack[c]{Pronoun and\\verbal focus}} \\
  & \textit{More / Less Frequent Verbs} & \\[0.5ex]

\midrule
\multirow{2}{*}{$\ell=7$}
  & \textit{With / Without Determiners} & \multirow{2}{*}{\shortstack[c]{Grammatical\\commitment}} \\
  & \textit{Certain / Uncertain} & \\

\bottomrule
\end{tabular}
}
\caption{
Representative sense-style alignments in iBERT-v3-10, based on sense activation (\autoref{tab:style-sense-activations}).\\All listed axes are the top-aligned style for their respective sense (i.e., highest probing activation).
}
\label{tab:sense-style-alignment}
\vspace{-1.3em}
\end{table}

\begin{table*}[t]
\centering\small
\resizebox{0.65\textwidth}{!}{

\begin{tabular}{c cccc ccc}
\toprule
\multirow{ 2}{*}{\textsc{Sense}}
& \multicolumn{4}{c}{\textsc{Target-Aligned Styles}} 
& \multicolumn{3}{c}{\textsc{Non-Target Styles}} \\
\cmidrule(lr){2-5} \cmidrule(lr){6-8}
& \textsc{\#Styles} & \textsc{Orig} & \textsc{Edit} & \textsc{$\Delta$Dist} (\%) 
& \textsc{Orig} & \textsc{Edit} & \textsc{$\Delta$Dist} (\%) \\
\midrule
$\ell=0$ & 3 & 0.812 & 0.459 & \textbf{\phantom{-}38.5} & 0.405 & 0.278 & \phantom{-} 1.2 \\
$\ell=1$ & 4 & 0.725 & 0.321 & \textbf{\phantom{-}58.0} & 0.390 & 0.250 & \phantom{-}15.2 \\
$\ell=2$ & 4 & 0.563 & 0.310 & \textbf{40.4} & 0.412 & 0.299 & \phantom{--}-2.2 \\
$\ell=3$ & 2 & 1.333 & 0.469 & \textbf{\phantom{-}65.7} & 0.414 & 0.324 & \phantom{-} -4.3 \\
$\ell=4$ & 2 & 0.336 & 0.168 & \textbf{\phantom{-}46.3} & 0.392 & 0.232 & \phantom{--} 7.7 \\
$\ell=5$ & 2 & 0.326 & 0.237 & \textbf{\phantom{-}26.1} & 0.452 & 0.271 & \phantom{--} 8.7 \\
$\ell=6$ & 2 & 0.531 & 0.395 & \textbf{\phantom{-}25.5} & 0.403 & 0.272 & -10.7 \\
$\ell=7$ & 2 & 0.353 & 0.328 & \textbf{\phantom{--} 7.0} & 0.488 & 0.347 & \phantom{-} -2.6 \\
\midrule
\end{tabular}
}
\vspace{-0.8em}
\caption{
Impact of sense-level ablation in iBERT-v3-10, using style groups from Table~\ref{tab:sense-style-alignment}. 
We report mean cosine distance to the opposite style centroid, before and after ablating sense $\ell$. 
Left: styles aligned with $\ell$; Right: all others. 
\textsc{Orig} = original; \textsc{Edit} = post-ablation; \textsc{$\Delta$Dist} = relative distance reduction (↑ = stronger disentanglement).
}
\label{tab:sense-ablation-summary}
\vspace{-0.6em}
\end{table*}

As highlighted before, the main goal of developing iBERT is to encode inputs as mixtures of interpretable sense vectors. To examine whether specific senses specialize in capturing coherent stylistic structure, we analyze the alignment between sense dimensions and styles via probing and controlled ablations. Table~\ref{tab:sense-style-alignment} groups the style pairs most strongly aligned with each sense for iBERT-v3-10, identified by first finding the highest activation senses per style, and then performing combinatorial optimization to group them (Appendix \autoref{tab:style-sense-activations}). Clear thematic clustering is observed: i.e., sense $\ell$=2 aligns with affective and expressive tone (sarcasm, metaphor), while $\ell$=0 captures surface-level markers like emoji use and personal pronouns. These patterns suggest that individual senses structurally specialize in distinct stylistic attributes.

To validate these groupings, \autoref{tab:sense-ablation-summary} reports the change in cosine distance between opposing style centroids before and after removing each sense contribution in final representation. When ablating a style-aligned sense, the separability between style polarities drops substantially (e.g., $66\%$ for $\ell=3$), confirming that key stylistic information is isolated within that sense. In contrast, most non-target styles exhibit low or negligible change in distance, indicating minimal collateral disruption: a hallmark of localized specialization.

Interestingly, for $\ell$=6 and $\ell$=7, ablating these senses harms separation among non-aligned styles (i.e., negative $\Delta$Dist), suggesting they may encode semantic content that generalizes across style boundaries. This aligns with the style groups they capture (e.g., pronoun and verb usage, determiners, certainty), which plausibly reflect broader discourse semantics. These observations highlight a nuanced balance between stylistic and semantic organization within iBERT's modular space.

Taken together, these results demonstrate that sense dimensions in iBERT naturally organize around interpretable stylistic clusters, enabling explicit, localized editing of style without disrupting unrelated attributes.

\subsection{Visualize Targeted Editing via Ablation}
\label{sec:edit}
\begin{figure*}[t]
    \centering

    \begin{subfigure}[t]{0.235\textwidth}
        \centering
        \includegraphics[width=\linewidth,trim={0.5cm 0.4cm 0.5cm 0.4cm},clip]
        {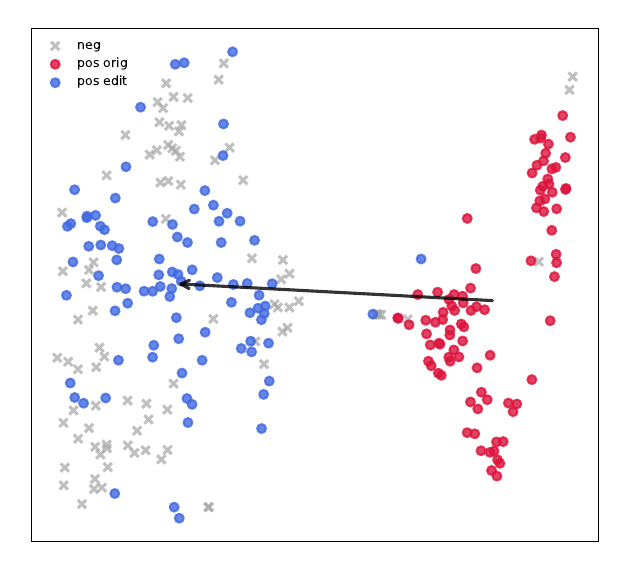}
        \caption{Text Emojis / No Emojis\\($\ell=1$, $\Delta$Dist: 84\,\%)}
        \label{fig:tsne_text_emojis}
    \end{subfigure}
    \hspace{0.26em}
    \begin{subfigure}[t]{0.235\textwidth}
        \centering
        \includegraphics[width=\linewidth,trim={0.5cm 0.4cm 0.5cm 0.4cm},clip]
        {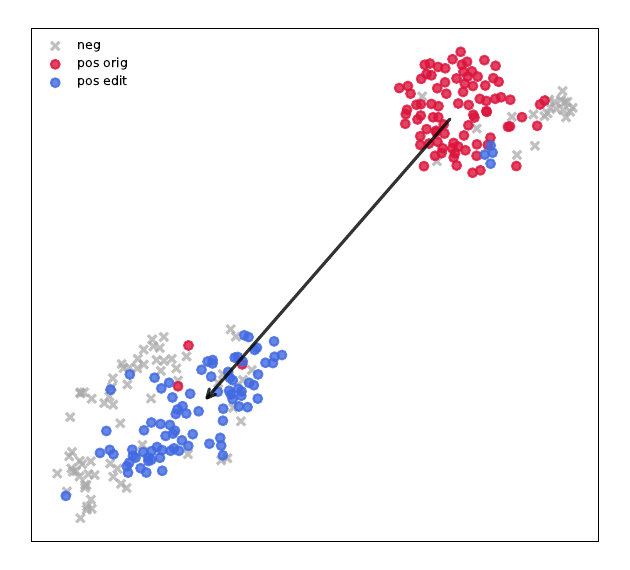}
        \caption{With/Without Nominalizations ($\ell$=4, $\Delta$Dist: 67\%)}
        \label{fig:tsne_contractions}
    \end{subfigure}
    \hspace{0.26em}
    \begin{subfigure}[t]{0.235\textwidth}
        \centering
        \includegraphics[width=\linewidth,trim={0.5cm 0.4cm 0.5cm 0.4cm},clip]
        {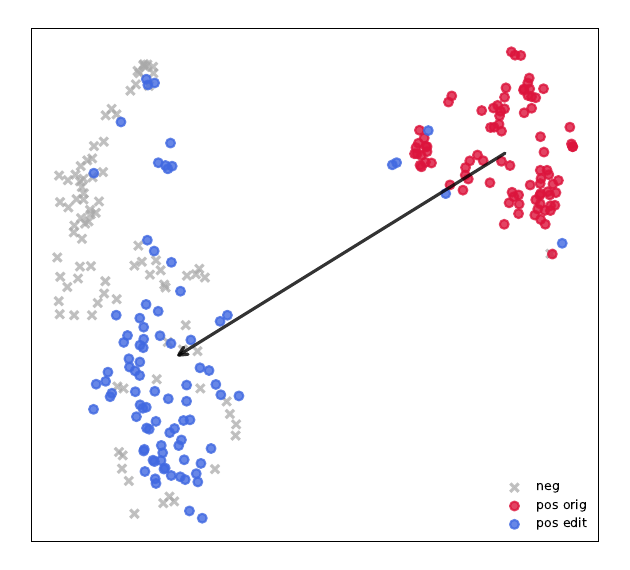}
        \caption{All Lowercase / Proper Caps ($\ell=7$, $\Delta$Dist: 70\%)}
        \label{fig:tsne_case}
    \end{subfigure}
    \vspace{-0.4em}
    \caption{
    Style-edit t-SNEs for iBERT-v3-10, ablating the most aligned sense $\ell$ for a given style. 
    Red: original positive samples; Blue: edited ablated positive samples; Gray: negative samples.
    Arrows are from original-positive-centroid, to: edited-positive-centroid. Edits control semantics along: 
    (a) visual form, (b) syntactic function, and (c) grammatical commitment. $\Delta$Dist: relative decrease in mean distance of positive samples to the negative centroid.
    }
    \label{fig:tsne-multi}
    \vspace{-0.9em}
\end{figure*}

To visualize iBERT's ability to localize and disentangle style, we conduct controlled ablations over its sense vectors. \autoref{fig:tsne-multi} shows t-SNE projections for three representative styles ablated along their most aligned senses (Table~\ref{tab:sense-style-alignment}). In all cases, ablating the target sense causes the positive samples (red) to move toward the negative centroid (gray), with relative distance dropping up to 84\%. These reflect localized edits validating iBERT's controllability.

\subsection{Visualize Non-Target Style Ablation}
\label{sec:appendix-tsne}
To ensure that iBERT's sense editing visualizations (\autoref{sec:edit}) are not generic artifacts of perturbation, \autoref{fig:tsne-non-target} ablates unrelated senses for unrelated (non-target) style contrasts.
In all cases, positive and negative samples remain well-separated, even as the positive cluster slightly shifts. This affirms that sense vectors act locally, without broad entanglement across unrelated styles.

Note that t-SNE primarily preserves local topology rather than absolute displacement; thus, these visualizations should be interpreted qualitatively. The clear contrast between targeted collapse (\autoref{fig:tsne-multi}) and stable separation in control (\autoref{fig:tsne-non-target}) supports our claims of modularity and editability.

\subsection{Quantifying Locality of Sense-Editing}
\label{sec:locality}

A core utility for controllable representations is \emph{locality}: editing a target attribute should only affect that attribute while minimally disturbing unrelated dimensions.
While Figure~\ref{fig:tsne-multi} illustrates targeted edits qualitatively, we now formalize this behavior.

For a given style $x$ (e.g., \emph{Metaphorical / Literal}), we identify the sense $\ell^*$ with the highest probing activation and ablate it during encoding by zeroing the corresponding sense gain. Let $P_x$ denote the positive examples for style $x$, and $P_x^{\text{orig}}$, $P_x^{\text{edit}}$ be their embeddings before and after ablating $\ell^*$, respectively. We compute cosine distance between a set of embeddings $A$ and a centroid $\mu$ as:
\begin{equation*}
d(A,\mu) = \frac{1}{|A|} \sum_{a \in A} \left(1 - \cos(a,\mu)\right).
\end{equation*}

\noindent
The \textit{target editability} is then:
\begin{equation*}
\Delta_x =
\frac{d(P_x^{\text{orig}}, \mu_x^{-}) - d(P_x^{\text{edit}}, \mu_x^{-})}{d(P_x^{\text{orig}}, \mu_x^{-})},
\end{equation*}
where $\mu_x^{-}$ is the centroid of negative examples for style $x$. 

\noindent
To measure \textit{non-target editing} (collateral effects), we evaluate how the edits for $x$ impact \emph{other} stylistic properties $y\neq x$, using $\mu_y^{-}$, which is the centroid of non-target $y$'s negative examples:
\begin{equation*}
\Delta_y =
\frac{d(P_x^{\text{orig}}, \mu_y^{-}) - d(P_x^{\text{edit}}, \mu_y^{-})}{d(P_x^{\text{orig}}, \mu_y^{-})}.
\end{equation*}
We report the following aggregate locality metrics:
\begin{equation*}
\begin{aligned}
\text{Avg.\;Other Shift} &= \frac{1}{|\{y \neq x\}|} \sum_{y \neq x} \Delta_y 
\label{eq:locality}
\end{aligned}
\end{equation*}
A useful and disentangled iBERT sense should ideally yield a large, positive $\Delta_x$ while keeping $\text{Avg.\;Other Shift}$ low, indicating minimal collateral influence on unrelated style attributes.

Table~\ref{tab:sense-ablation-summary} shows that ablating a style-aligned sense yields large reductions in target separability (up to 65\%), while inducing minimal average shift on non-target styles.
This confirms that stylistic attributes are concentrated in specific sense dimensions rather than distributed diffusely across the embedding.

\begin{figure}[t]
    \centering

    \begin{subfigure}[t]{0.23\textwidth}
        \centering
        \includegraphics[width=\linewidth]{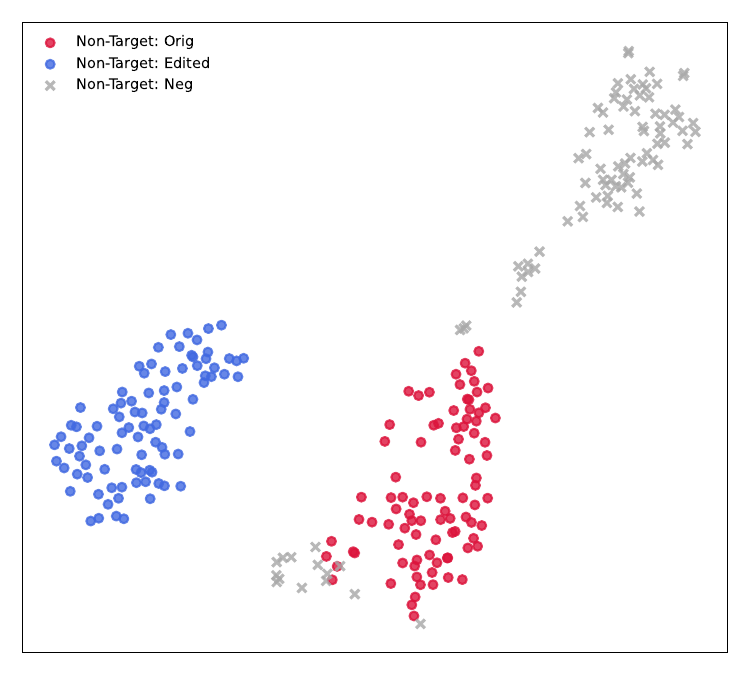}
        \caption{$\ell=0$ ablated that aligns with More/Less conjunctions}
        \label{fig:tsne_text_emojis}
    \end{subfigure}
    \hfill
    \begin{subfigure}[t]{0.23\textwidth}
        \centering
        \includegraphics[width=\linewidth]{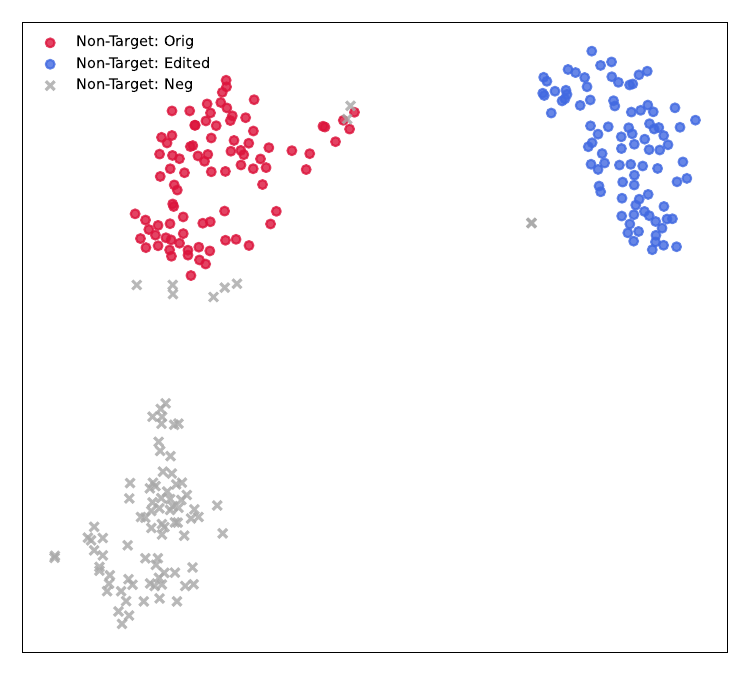}
        \caption{$\ell=2$ ablated that aligns with Sarcasm / No sarcasm}
        \label{fig:tsne_case}
    \end{subfigure}
    \caption{All Upper Case/Proper Capitalization style ($\ell^*$=1) maintains separation on ablating other senses ($\ell$=\{0, 2\}). 
    Colors as in Fig.~\ref{fig:tsne-multi}- Red: original, Blue: edited, Gray: negatives.
    }
    \label{fig:tsne-non-target}
\end{figure}

As an added control, we refer again to Figure~\ref{fig:tsne-non-target}, which shows that ablating a non-primary sense (i.e., not the top sense $\ell^*$) for a given style leaves its attribute structures intact. This supports the view that iBERT's sense representations are modular, with sense edits affecting the intended attributes, rather than causing broad changes across the space.

\section{Discussion}
\label{sec:discussion}

\paragraph{Interpretability without performance tradeoff.}
iBERT delivers inherently interpretable embeddings, while maintaining competitive performance across both style-intensive and mixed-style tasks. Because each input is represented as a sparse mixture of sense vectors, users can inspect, edit, or ablate specific dimensions with explicit attribution to stylistic or semantic features. Unlike post-hoc attribution methods, these controls are embedded in model's structure that directly alters representation.

\paragraph{Disentanglement through structure and signal.}
Our results confirm that architectural inductive bias alone is insufficient: meaningful disentanglement emerges only with high-quality supervision (e.g., SynthSTEL). Under proxy-style signals (e.g., WG-only), sense specialization weakens, and style polarity collapses. This co-dependence of structure and supervision underscores a key design insight: 
iBERT is not style-specific, but learns to modularize whichever axes are discriminative in the data.

\paragraph{Localized edits, global control.}
Ablating a single sense (e.g., $\ell{=}1$ for emoji, or $\ell{=}4$ for nominalization) shifts the representation toward the negative style without disrupting unrelated styles. These transformations are visible both numerically (Table~\ref{tab:sense-ablation-summary}) and spatially (Fig.~\ref{fig:tsne-multi}), confirming that iBERT learns axis-aligned subspaces that can be controlled with precision. This kind of intervention is infeasible in dense encoders like SBERT or BERT.

\paragraph{Semantic resilience in modular space.}
Despite modularizing stylistic signals, iBERT retains generalization capacity in semantically entangled tasks (e.g., PAN13/15). Even sense vectors associated with stylistic axes (e.g., $\ell{=}6,7$) appear to encode semantic scaffolding, since ablating them harms non-target style separability. This indicates that the modularity does not rigidly isolate content, but supports a soft balance between style and semantics.

\paragraph{Bridging representation learning and sociolinguistics.}
Emergent sense specializations align with sociolinguistic constructs such as formality, register, modality, and expressive tone. This alignment arises without explicit annotation, suggesting iBERT's decomposition mirrors meaningful axes of human communication. These findings open pathways for interpretable modeling in domains like forensics, social analysis, and fairness auditing.

\paragraph{Applications.}
iBERT is not a style model, but a general architecture to tease out discriminative signals into interpretable senses, and it supports a range of feature—conditioned retrieval, classifier debiasing via sense ablation, latent data augmentation, and embedding-level control in RAG pipelines. See Appendix~\ref{sec:appendix-broader-applications} for more details.

\section{Conclusion}
\label{sec:conclusion}

We introduced \textbf{iBERT} (interpretable-BERT), a modular encoder that produces sparse, interpretable, and controllable embeddings without compromising performance. By structuring each input as a mixture over reusable sense vectors, iBERT supports inspection, editing, and attribution of linguistic properties within the representation.

iBERT matches or outperforms dense baselines  across stylistic and semantic tasks like STEL, SoC, and PAN, while enabling: (a) disentangled encoding of style and semantics via sense-probing, (b) precise edits (e.g., formality), and (c) alignment of emergent senses with sociolinguistic axes.

Rather than relying on post-hoc interpretability methods, iBERT is designed to be \textit{explainable by construction}. Its structure reveals how representations evolve and interact—making it a viable backbone for transparent, modular NLP.

We release our models to encourage further exploration of compositional embeddings, zero-shot control, and sense-aware generation. \mbox{iBERT} offers a step toward embedding models that are not only powerful, but inherently understandable.

\section*{Model and Code} We release the
models, demonstration, and code for other researchers to use. These are made available at \href{https://github.com/vishalanand/iBERT}{https://github.com/vishalanand/iBERT} and  \href{https://iBERT.io}{https://iBERT.io} .

\vfill

\section*{Limitations}
\label{sec:limitations}

While iBERT enables modular, interpretable representations with strong empirical performance, a few limitations remain. The model’s design introduces a tradeoff between disentangled structure and tight semantic coupling: in tasks requiring deep contextual integration, this can lead to slight performance gaps compared to dense encoders.

In addition, sense specialization can be sensitive to sequence length. Our training datasets: SynthSTEL (median 19 tokens) and WG (24 tokens) offer limited context, which may constrain the model's ability to fully activate or separate sense subspaces. This could partially explain weaker gains in tasks involving longer-form semantics.

Lastly, while we focused on English, extending sense-level modularity to morphologically rich or low-resource languages remains an open direction.

These limitations offer promising opportunities to expand iBERT into a broader, language-agnostic foundation for interpretable and controllable representation learning.

\section*{Ethics Statement}

iBERT is developed as a general-purpose encoder that produces interpretable, controllable sentence representations by decomposing inputs into structured mixtures over sense vectors. This modular design supports greater transparency in how linguistic features—whether semantic, syntactic, or stylistic—contribute to downstream decisions.

We believe such structured interpretability fosters accountability in language technologies. However, we recognize potential misuse: modular representations that expose linguistic traits may be leveraged for tasks like profiling or authorship attribution, which can carry privacy risks in sensitive settings. These risks are pronounced when models are applied without consent, or outside the distribution of the data used during training.

iBERT is trained on public data, including synthetic edits (StyleSynth) and large-scale web corpora. As with all pretrained models, underlying biases in this data may influence the dimensions that emerge. We encourage careful evaluation before deployment in applications with societal or demographic impact.

We release our models and code for research use, with the goal of encouraging safe, transparent, and interpretable alternatives to opaque neural representations.

\section*{Acknowledgments}
\label{sec:acknowledgements}
We thank John Hewitt (Columbia University) for helpful discussions during experiment and model design. We thank \mbox{Dr. Cheng Wu (\mbox{Microsoft}) for} his insights during modeling and experiment design, and supporting our research. We thank Ajay Patel (University of Pennsylvania), Anna Wegmann (Utrecht University), Nicholas Andrews (Johns Hopkins University), Joel Tetreault (\mbox{Dataminr}), and Sudha Rao (Microsoft Research) for granting access to relevant datasets. We also thank David M. Rothschild (Microsoft Research), Vishal \mbox{Chowdhary} (Microsoft Office AI), Yasaman Ameri (Visa Inc.), and Aparna Balagopalan (MIT) for early conversations that helped shape the first author's thinking around research problem formulation and scoping.

\appendix

\section{Appendix}
\label{sec:appendix}

\begin{table*}[t]
\centering \footnotesize
\setlength{\tabcolsep}{5pt}
\resizebox{\textwidth}{!}{
\begin{tabular}{ll | c c c c c | c c c c c}
\toprule
& \multirow{2.5}{*}{\textsc{Models}} & \multicolumn{5}{c|}{\textsc{128 tokens}} & \multicolumn{5}{c}{\textsc{512 tokens}} \\
\cmidrule(lr){3-7} \cmidrule(lr){8-12}
\textbf{} & & \cellcolor{gray!12}\textsc{Avg} & \textsc{PAN11} & \textsc{PAN13} & \textsc{PAN14} & \textsc{PAN15} & \cellcolor{gray!12}\textsc{Avg} & \textsc{PAN11} & \textsc{PAN13} & \textsc{PAN14} & \textsc{PAN15} \\
\midrule
\multirow{5}{*}{\rotatebox{90}{\small \textsc{SD-only}}}
& BERT & \cellcolor{gray!12} \textit{60.57} & 79.34 & \textbf{56.61} & \textbf{52.84} & \textbf{\underline{53.47}} & \cellcolor{gray!12} \textbf{\underline{63.52}} & \textit{78.48} & \textbf{69.89} & \textbf{56.40} & 49.31 \\
& iBERT-v1 & \cellcolor{gray!12} \textbf{60.88} & \textbf{\underline{87.64}} & 53.50 & 51.37 & 50.99 & \cellcolor{gray!12} 59.49 & \textbf{\underline{79.19}} & 55.95 & 52.52 & 50.31 \\
& iBERT-v2 & \cellcolor{gray!12} 58.95 & 83.49 & 48.70 & 51.03 & \textit{52.57} & \cellcolor{gray!12} 59.41 & 72.83 & 57.88 & 52.86 & \textbf{54.08} \\
& iBERT-v3-1 & \cellcolor{gray!12} 58.15 & 80.42 & 50.56 & 50.98 & 50.64 & \cellcolor{gray!12} 59.39 & \textit{78.54} & 57.04 & 52.60 & 49.37 \\
& iBERT-v3-10 & \cellcolor{gray!12} 58.29 & 80.73 & 49.61 & 50.47 & 52.36 & \cellcolor{gray!12} 59.41 & 77.30 & 56.19 & 53.62 & 50.51 \\
\midrule
\multirow{5}{*}{\rotatebox{90}{\small \textsc{WG+SD}}}
& BERT & \cellcolor{gray!12} 58.96 & 68.82 & 63.04 & 52.55 & \textbf{51.43} & \cellcolor{gray!12} \textbf{62.76} & 70.35 & \textbf{\underline{71.46}} & \textbf{57.00} & \textit{52.23} \\
& iBERT-v1 & \cellcolor{gray!12} 58.96 & 63.75 & \textbf{66.87} & \textbf{54.45} & \textit{50.77} & \cellcolor{gray!12} 61.53 & \textbf{71.64} & 67.92 & 55.91 & 50.64 \\
& iBERT-v2 & \cellcolor{gray!12} 58.47 & 70.25 & 59.64 & 53.10 & \textit{50.88} & \cellcolor{gray!12} 59.54 & 66.60 & 65.32 & 54.75 & 51.50 \\
& iBERT-v3-1 & \cellcolor{gray!12} 58.56 & 69.08 & 61.33 & \textit{53.90} & 49.92 & \cellcolor{gray!12} 61.43 & 70.01 & 66.15 & \textit{56.82} & \textbf{52.72} \\
& iBERT-v3-10 & \cellcolor{gray!12} \textbf{\underline{60.96}} & \textbf{79.02} & 60.72 & 52.91 & \textit{51.19} & \cellcolor{gray!12} 60.34 & \textit{71.03} & 63.72 & 55.95 & 50.65 \\
\midrule
\multirow{5}{*}{\rotatebox{90}{\small \textsc{WG-only}}}
& BERT & \cellcolor{gray!12} 58.20 & 65.59 & 62.03 & 53.55 & \textbf{51.62} & \cellcolor{gray!12} \textbf{61.92} & 65.05 & \textbf{70.18} & \textit{58.19} & \textbf{\underline{54.27}} \\
& iBERT-v1 & \cellcolor{gray!12} \textit{59.29} & 64.69 & \textbf{\underline{67.99}} & \textbf{\underline{55.80}} & 48.69 & \cellcolor{gray!12} 60.74 & \textit{67.14} & 66.89 & \textit{57.77} & 51.17 \\
& iBERT-v2 & \cellcolor{gray!12} 57.77 & 66.93 & 62.27 & 54.17 & 47.74 & \cellcolor{gray!12} 58.63 & 62.41 & 65.75 & 56.58 & 49.78 \\
& iBERT-v3-1 & \cellcolor{gray!12} \textit{59.47} & 66.12 & 66.98 & \textit{55.68} & 49.11 & \cellcolor{gray!12} 60.87 & 65.92 & 68.87 & \textbf{\underline{58.29}} & 50.38 \\
& iBERT-v3-10 & \cellcolor{gray!12} \textbf{59.87} & \textbf{69.70} & 65.15 & \textit{55.23} & 49.40 & \cellcolor{gray!12} \textit{61.07} & \textbf{68.02} & 67.73 & \textit{57.92} & 50.59 \\
\midrule
\bottomrule
\end{tabular}
}
\caption{AUC (\%) on PAN 11/13/14/15 authorship verification tasks for \textbf{128 vs. 512 token-size models}. Rows are grouped by training setup (\textbf{SD}, \textbf{WG}, \textbf{WG+SD}). \textbf{Bold} indicates the best mean per group-token combination. \underline{Underline} highlights global best per token. \textit{Italics} mark values reasonably close to the best per group. }
\label{tab:pan_sd_wg_128_512}
\end{table*}

\begin{figure}[t]
  \centering

  \begin{subfigure}{0.23\textwidth}
    \includegraphics[width=\linewidth]{"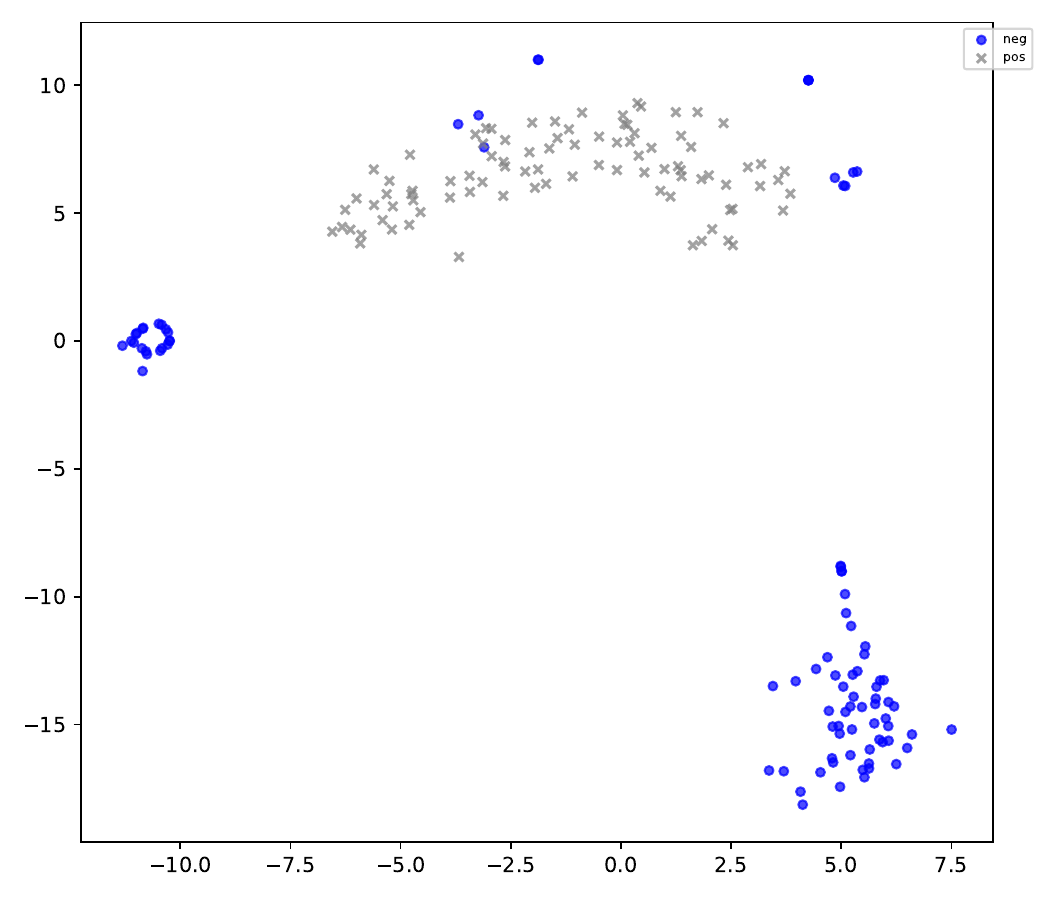"}
    \caption{BERT:\\Sentences with Misspellings}
    \label{fig:tsne-vanilla-numsub}
  \end{subfigure}
  \hfill
  \begin{subfigure}{0.23\textwidth}
    \includegraphics[width=\linewidth]{"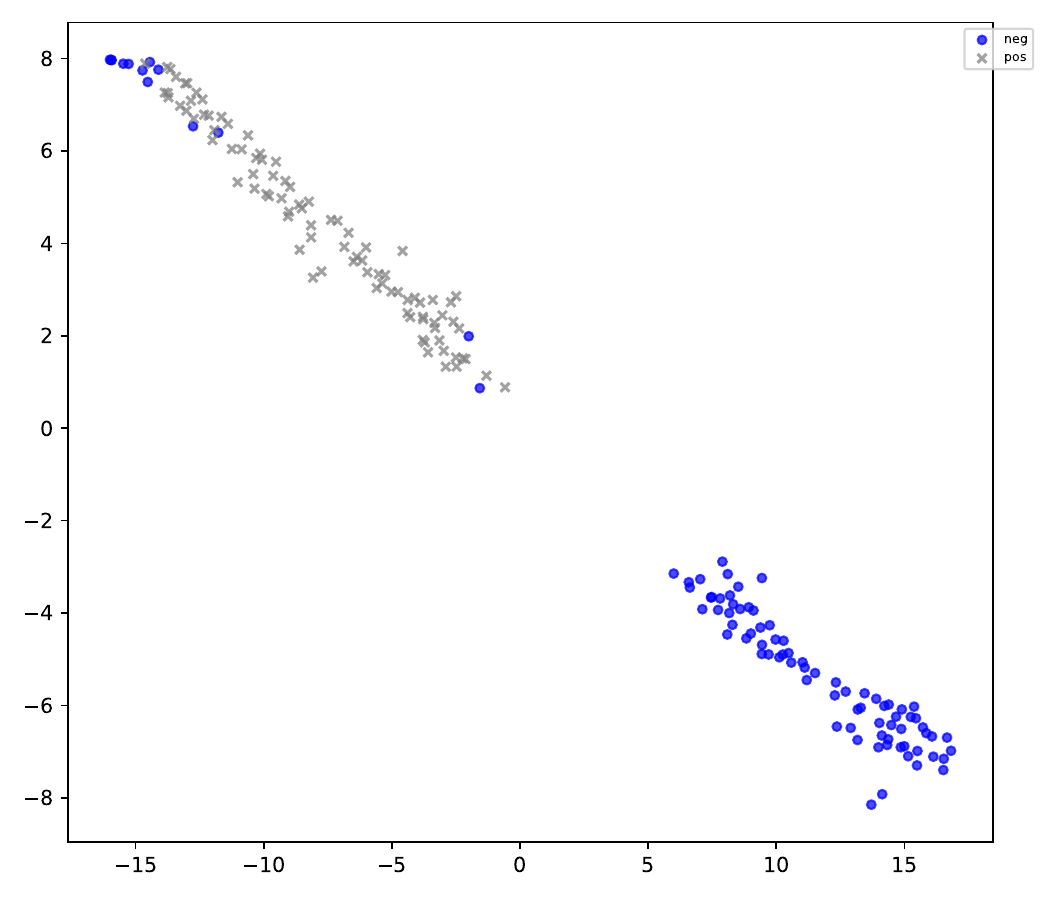"}
    \caption{iBERT-v2: \\Sentences with Misspellings}
    \label{fig:tsne-iBERT-numsub}
  \end{subfigure}

  \vspace{1em}

  \begin{subfigure}{0.23\textwidth}
    \includegraphics[width=\linewidth]{"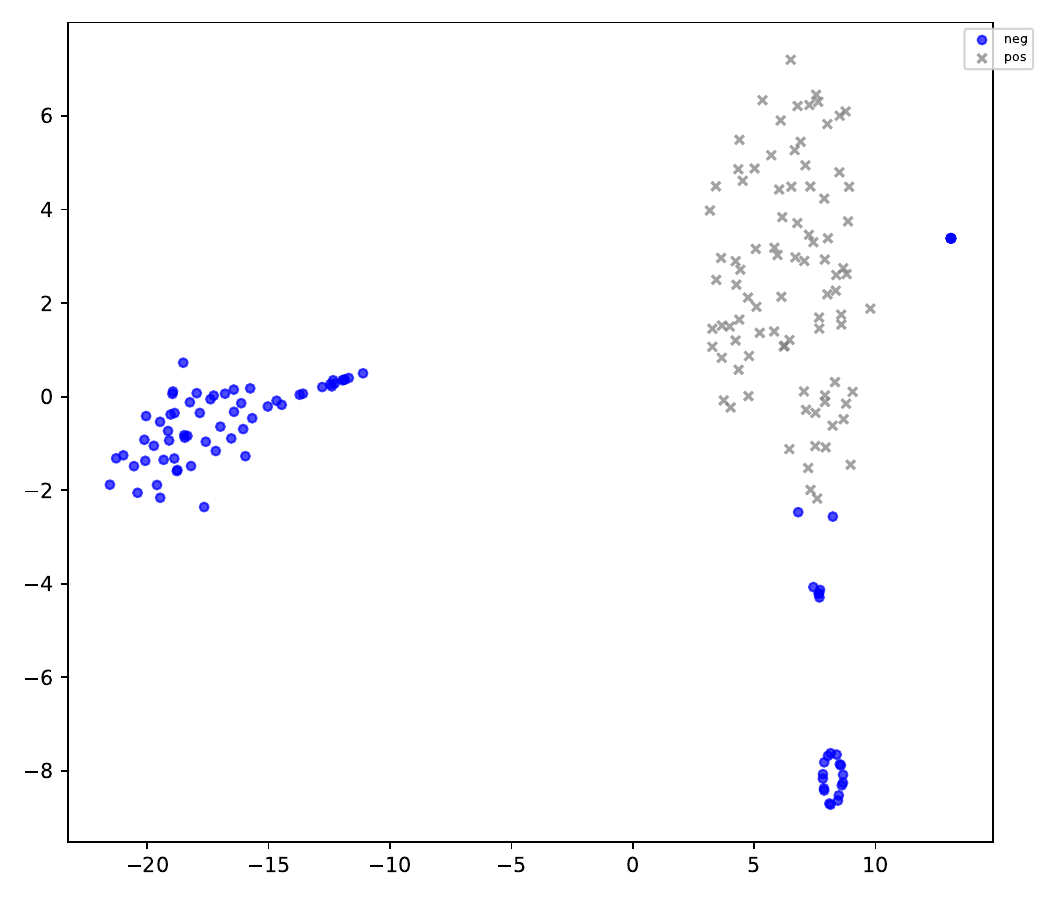"}
    \caption{BERT: \\Emoji Usage}
    \label{fig:tsne-vanilla-pronoun}
  \end{subfigure}
  \hfill
  \begin{subfigure}{0.23\textwidth}
    \includegraphics[width=\linewidth]{"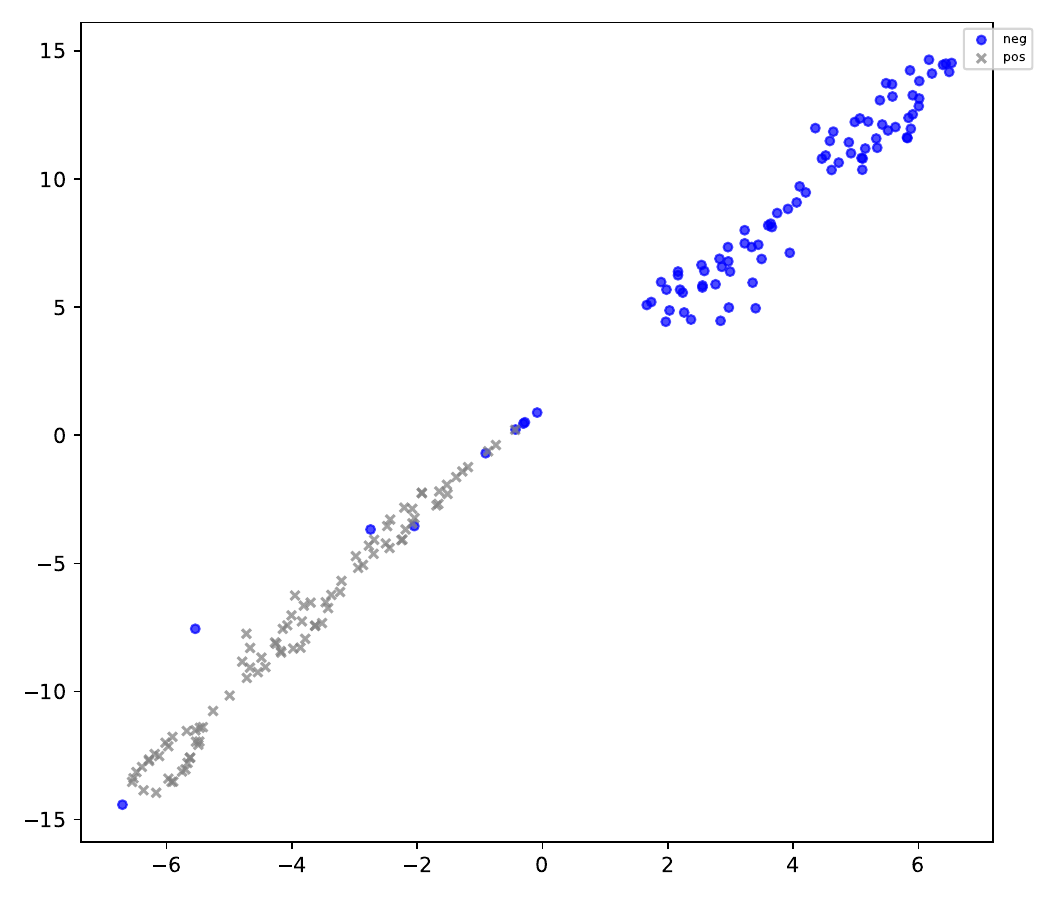"}
    \caption{iBERT-v2: \\Emoji Usage}
    \label{fig:tsne-iBERT-pronoun}
  \end{subfigure}

  \caption{t-SNE projections of sentence embeddings for SynthSTEL. iBERT consistently separates contrastive style variants (e.g., misspelled sentences, emoji usage) better than BERT, showing clearer margins and lower entanglement. Blue points represent positive samples and gray crosses represent negative samples. Despite iBERT-v2 being the most underperforming iBERT variant it still matches BERT's performance, and the separation of positive and negative styles is cleaner.}
  \label{fig:appendix-tsne}
\end{figure}

\begin{figure*}[t]
    \centering
    \begin{subfigure}{\textwidth}
      \centering
      \includegraphics[width=\linewidth,trim={0.0cm 0.0cm 0.0cm 0.08cm},clip]{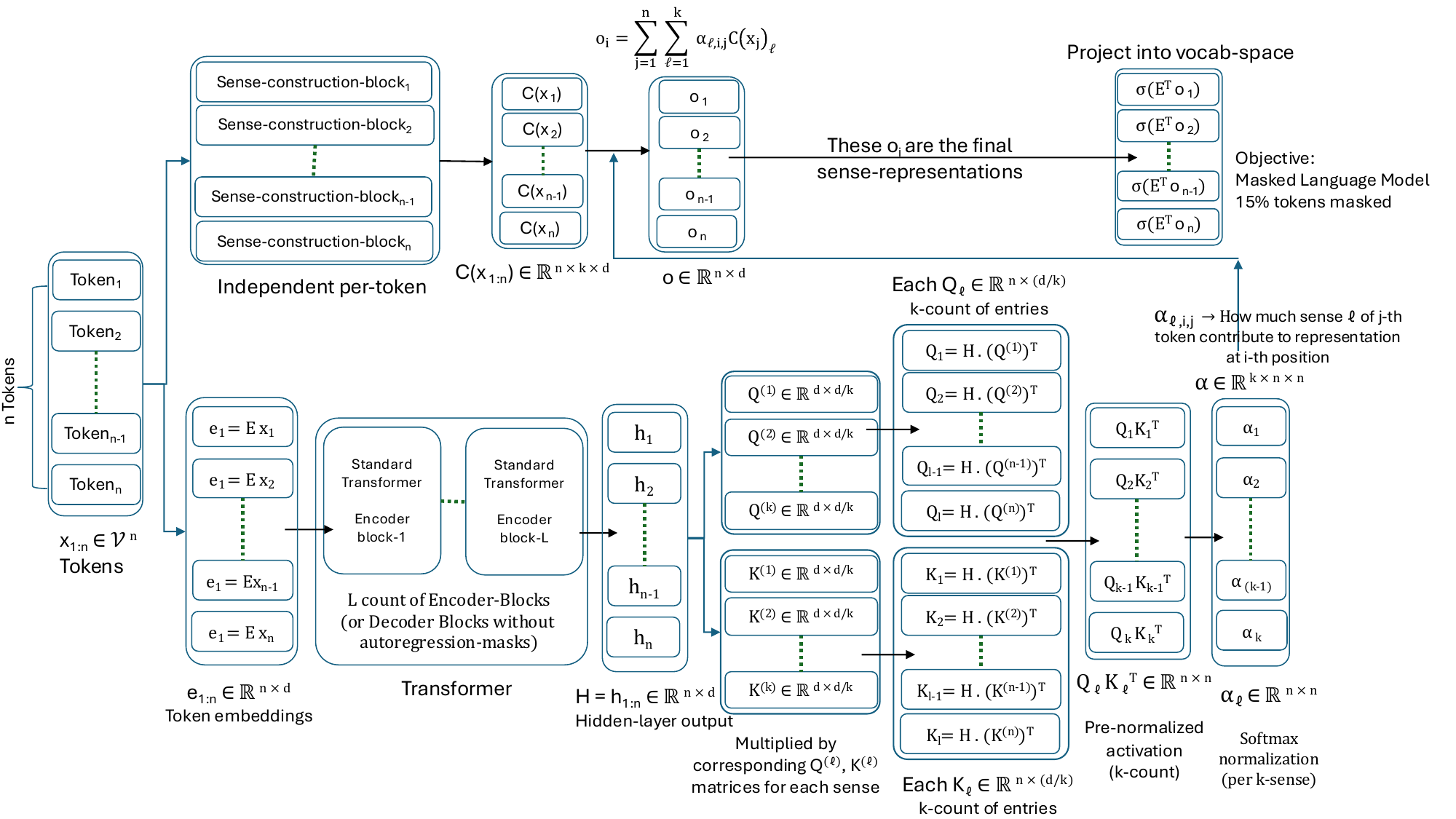}
      \caption{iBERT (MLM) overview (\S\ref{sec:method}). Each input token is projected to $k$ sense vectors, contextualized using attention-weighted combinations across all positions. This is a drop-in replacement of BERT for easy adoption.}
      \vspace{1.0em}
      \label{fig:iBERT-MLM}
    \end{subfigure}
    \hfill
    \begin{subfigure}[t]{0.45\textwidth}
        \centering
        \includegraphics[height=5cm, keepaspectratio,trim={1.14cm 1.26cm 15.7cm 3.3cm},clip]{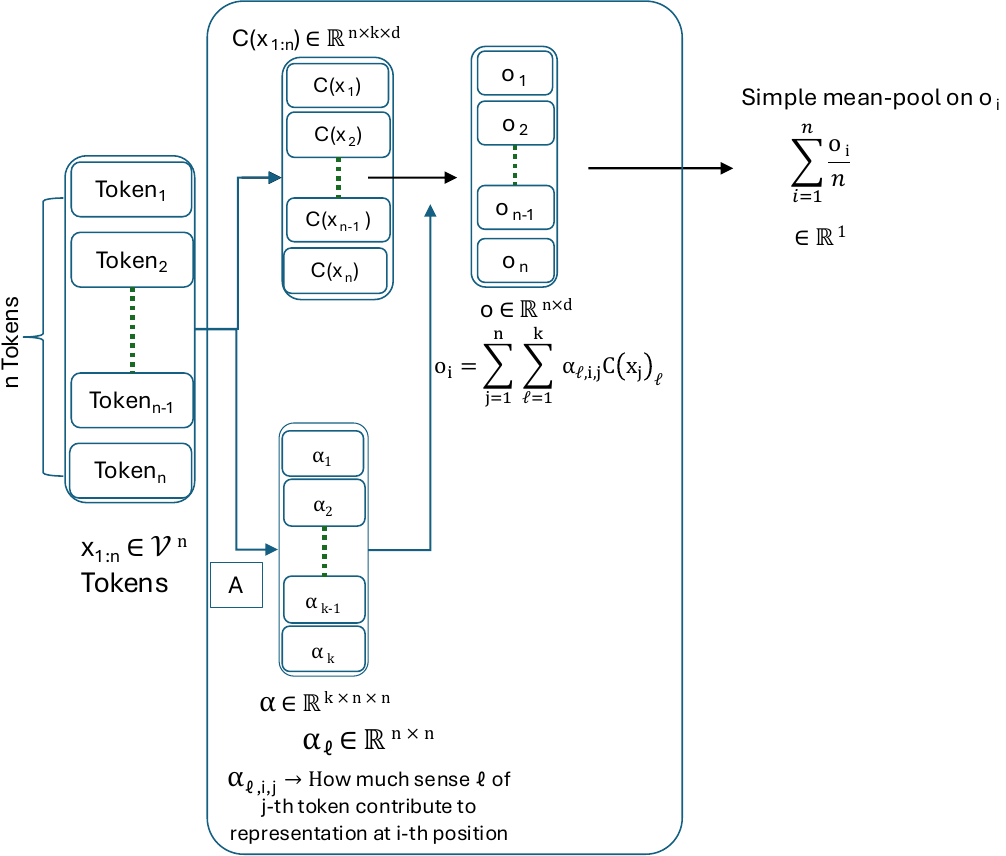}
        \caption{iBERT-v1: simple mean pooling over token representations $o_i$.}
      \label{fig:iBERT-v1}
    \end{subfigure}
    \hfill
    \begin{subfigure}[t]{0.45\textwidth}
        \centering
        \includegraphics[height=5cm, keepaspectratio,trim={1.14cm 1.26cm 13.0cm 3.3cm},clip]{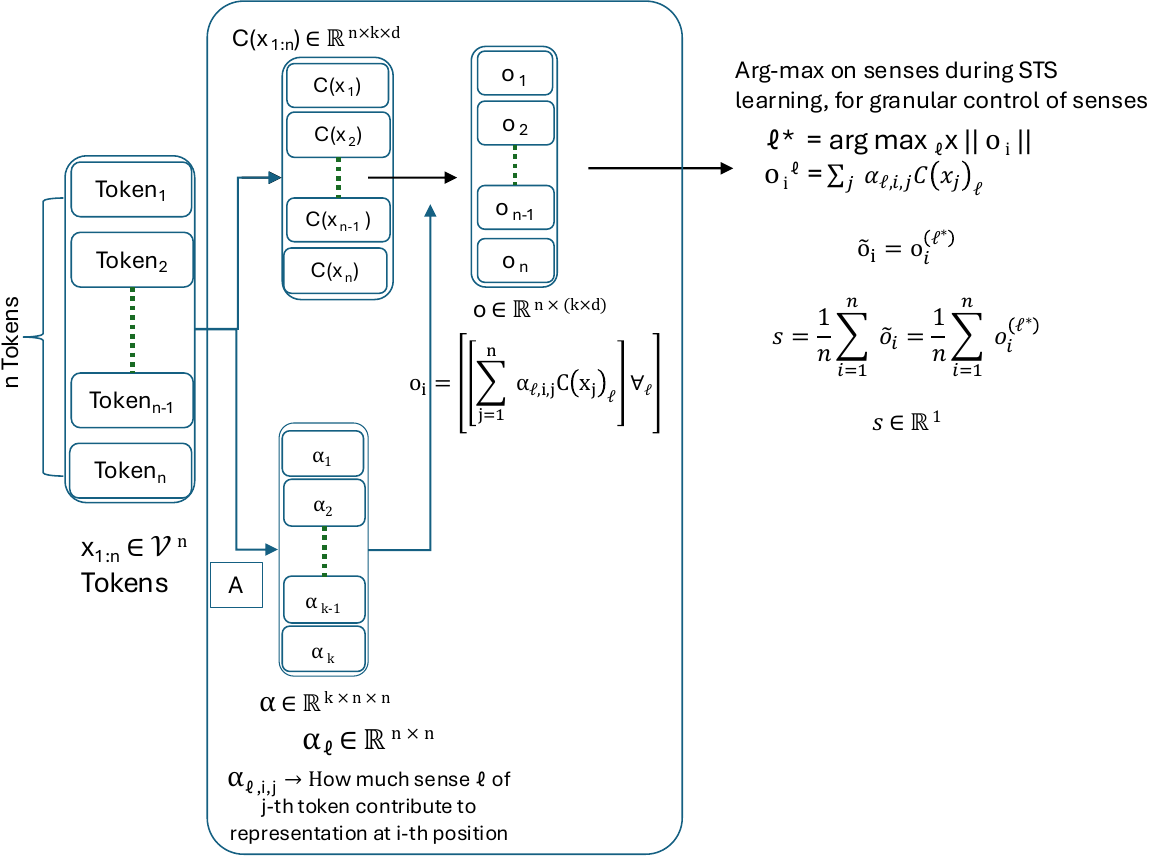}
        \caption{iBERT-v2: a single dominant sense index $\ell^*$ is selected, and only the corresponding components $o_i^{(\ell^*)}$ are pooled.}
        \vspace{1.0em}
      \label{fig:iBERT-v2}
    \end{subfigure}
    \hfill
    \begin{subfigure}[t]{0.45\textwidth}
        \centering
        \includegraphics[width=\linewidth, trim={2.29cm 10.0cm 14.8cm 1.59cm},clip]{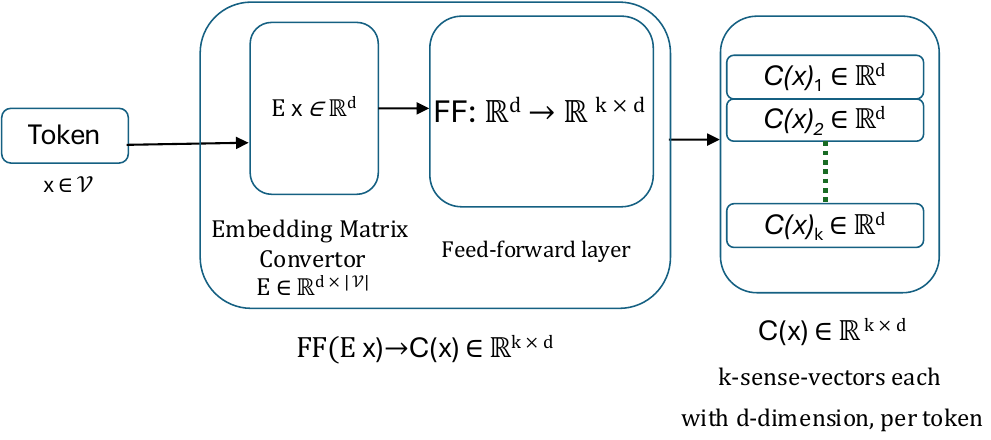}
        \caption{Sense Construction Block: each token is mapped via a feedforward layer to \(k\) sense vectors of dimension \(d\).}
      \label{fig:sense-construction}
    \end{subfigure}
    \caption{(a) shows the detailed technical iBERT architecture; (b–c) illustrate two pooling strategies used in sentence encoding (v1 and v2). The softmax-blend pooling variant (v3) lies between these and is described in Section ~\autoref{subsection:iBERT-sentence-embeddings}. (d) details the sense construction block.}
    \label{fig:iBERT-detailed}
\end{figure*}

\begin{table}[t]
\centering\small
\begin{tabular}{ll lcc}
\toprule
\textbf{Model} & \textbf{$\tau$} & Aggregation & STEL$\uparrow$ & SoC$\uparrow$ \\
\midrule
v1 & $\infty$ & Mean (uniform blend) & 37.5 & 92.3 \\
v3-10 & 10       & Peaked blend         & 38.1 & \textbf{92.8} \\
v3-1 & 1        & Softmax blend        & \textbf{38.3} & \textbf{92.8} \\
v2 & 0        & Top-sense only       & 28.5 & 88.7 \\
\bottomrule
\end{tabular}
\caption{Effect of aggregation sharpness $\tau$ on iBERT, trained on StyleSynth triplets (128 token size).}
\label{tab:tau}
\end{table}
\begin{table*}[t]
\centering
\resizebox{0.98\textwidth}{!}{
\begin{tabular}{lccccccccc}
\toprule
\textsc{Style} & \textsc{$\ell=0$} & \textsc{$\ell=1$} & \textsc{$\ell=2$} & \textsc{$\ell=3$} & \textsc{$\ell=4$} & \textsc{$\ell=5$} & \textsc{$\ell=6$} & \textsc{$\ell=7$} & \textsc{Best} \\
\midrule
Active / Passive & 0.6222 & 0.6222 & 0.5278 & 0.6278 & 0.6111 & 0.6444 & 0.5944 & 0.6333 & 5 \\
Affective process / Perceptual process & 0.6389 & 0.5444 & 0.6500 & 0.5889 & 0.6222 & 0.5833 & 0.6444 & 0.6278 & 2 \\
Affective processes / Cognitive processes & 0.6722 & 0.5833 & 0.6833 & 0.6111 & 0.6833 & 0.6389 & 0.7000 & 0.6722 & 6 \\
All Lower Case / Proper Capitalization & 0.9778 & 0.7611 & 0.9667 & 0.8444 & 0.9667 & 0.8778 & 0.9833 & 0.9833 & 7 \\
All Upper Case / Proper Capitalization & 0.9278 & 0.9444 & 0.9278 & 0.9333 & 0.9333 & 0.9389 & 0.9167 & 0.9222 & 1 \\
Certain / Uncertain & 0.6444 & 0.6056 & 0.6167 & 0.6444 & 0.6389 & 0.6500 & 0.6833 & 0.6944 & 7 \\
Cognitive process / Perceptual process & 0.6444 & 0.6167 & 0.5500 & 0.6500 & 0.6500 & 0.6500 & 0.6444 & 0.6667 & 7 \\
Complex / Simple & 0.5944 & 0.5444 & 0.6278 & 0.5611 & 0.6000 & 0.5556 & 0.6333 & 0.5944 & 6 \\
Fluent sentence / Disfluent sentence & 0.6278 & 0.5889 & 0.5111 & 0.6167 & 0.6111 & 0.6167 & 0.6444 & 0.6000 & 6 \\
Formal / Informal & 0.7833 & 0.7389 & 0.8167 & 0.7667 & 0.8167 & 0.7667 & 0.7667 & 0.7778 & 2 \\
Long average word length / Short average word length & 0.8389 & 0.8444 & 0.7944 & 0.8444 & 0.8389 & 0.8333 & 0.8278 & 0.8389 & 1 \\
Offensive / Non-Offensive & 0.9111 & 0.7389 & 0.9444 & 0.7611 & 0.9056 & 0.7722 & 0.9111 & 0.9000 & 2 \\
Polite / Impolite & 0.6389 & 0.4167 & 0.7333 & 0.4833 & 0.6667 & 0.4889 & 0.6444 & 0.5833 & 2 \\
Positive / Negative & 0.5944 & 0.5111 & 0.6944 & 0.5222 & 0.5944 & 0.5111 & 0.6056 & 0.5556 & 2 \\
Present-focused / Future-focused & 0.7000 & 0.6278 & 0.5556 & 0.6556 & 0.6833 & 0.6556 & 0.6778 & 0.7000 & 7 \\
Present-focused / Past-focused & 0.6778 & 0.5778 & 0.6333 & 0.5944 & 0.6278 & 0.5833 & 0.6944 & 0.6667 & 6 \\
Self-focused / Audience-focused & 0.8111 & 0.5778 & 0.7944 & 0.6444 & 0.7889 & 0.6389 & 0.8278 & 0.8111 & 6 \\
Self-focused / Inclusive-focused & 0.7500 & 0.5222 & 0.7556 & 0.5611 & 0.7000 & 0.5667 & 0.7333 & 0.7222 & 2 \\
Self-focused / Third-person singular & 0.6333 & 0.6056 & 0.5833 & 0.6167 & 0.6167 & 0.6111 & 0.6222 & 0.6444 & 7 \\
Self-focused / You-focused & 0.7944 & 0.5500 & 0.7778 & 0.6278 & 0.7556 & 0.6333 & 0.7889 & 0.7667 & 0 \\
Sentence With a Few Misspelled Words / Normal Sentence & 0.8889 & 0.8722 & 0.8389 & 0.8889 & 0.8944 & 0.8889 & 0.9111 & 0.9056 & 6 \\
Text Emojis / No Emojis & 0.9278 & 0.9333 & 0.9167 & 0.9278 & 0.9333 & 0.9333 & 0.9167 & 0.9167 & 1 \\
With Emojis / No Emojis & 0.9389 & 0.9389 & 0.9222 & 0.9389 & 0.9389 & 0.9389 & 0.9278 & 0.9389 & 0 \\
With Humor / Without Humor & 0.7556 & 0.6944 & 0.8000 & 0.7167 & 0.7667 & 0.7167 & 0.7778 & 0.7222 & 2 \\
With articles / Less frequent articles & 0.6889 & 0.6889 & 0.6000 & 0.6889 & 0.6833 & 0.6833 & 0.7278 & 0.7167 & 6 \\
With common verbs / Less frequent common verbs & 0.7889 & 0.7278 & 0.7611 & 0.7111 & 0.7444 & 0.7111 & 0.8111 & 0.7444 & 6 \\
With conjunctions / Less frequent conjunctions & 0.7944 & 0.7667 & 0.7944 & 0.7500 & 0.7833 & 0.7500 & 0.7833 & 0.7444 & 0 \\
With contractions / Without contractions & 0.8167 & 0.5778 & 0.7056 & 0.6722 & 0.7722 & 0.7000 & 0.8333 & 0.8278 & 6 \\
With determiners / Less frequent determiners & 0.6778 & 0.7222 & 0.6389 & 0.7056 & 0.6889 & 0.7000 & 0.7167 & 0.7278 & 7 \\
With digits / Less frequent digits & 0.8889 & 0.8389 & 0.8889 & 0.8556 & 0.8833 & 0.8667 & 0.8667 & 0.8778 & 0 \\
With frequent punctuation / Less Frequent punctuation & 0.6667 & 0.6167 & 0.7111 & 0.6389 & 0.6778 & 0.6389 & 0.7278 & 0.7000 & 6 \\
With function words / Less frequent function words & 0.7056 & 0.6944 & 0.6944 & 0.7056 & 0.7111 & 0.7056 & 0.7056 & 0.6944 & 4 \\
With metaphor / Without metaphor & 0.7278 & 0.6889 & 0.8222 & 0.6944 & 0.7667 & 0.7056 & 0.7389 & 0.7222 & 2 \\
With nominalizations / Without nominalizations & 0.8167 & 0.8056 & 0.8222 & 0.8111 & 0.8222 & 0.8222 & 0.7889 & 0.8000 & 4 \\
With number substitution / Without number substitution & 0.9278 & 0.9111 & 0.9389 & 0.9167 & 0.9222 & 0.9167 & 0.9222 & 0.9167 & 2 \\
With personal pronouns / Less frequent pronouns & 0.7389 & 0.7389 & 0.7111 & 0.7111 & 0.7167 & 0.7056 & 0.7222 & 0.7000 & 0 \\
With prepositions / Less frequent prepositions & 0.6333 & 0.6722 & 0.6111 & 0.6889 & 0.6611 & 0.6833 & 0.6667 & 0.6778 & 3 \\
With pronouns / Less frequent pronouns & 0.7611 & 0.7222 & 0.7444 & 0.7111 & 0.7278 & 0.7056 & 0.7889 & 0.7222 & 6 \\
With sarcasm / Without sarcasm & 0.7778 & 0.7278 & 0.8167 & 0.7278 & 0.7889 & 0.7333 & 0.7833 & 0.7278 & 2 \\
With uppercase letters / Without uppercase letters & 0.6556 & 0.5333 & 0.6722 & 0.5444 & 0.6611 & 0.5444 & 0.6500 & 0.6278 & 2 \\
\bottomrule
\end{tabular}
}
\caption{[iBERT-v3-10 (128-tokens)] Probing activations for style classes across all senses ($\ell = 0$ to $7$). The \textsc{Best} column indicates the latent sense with the highest activation for each style. Style names in this table reflect the original labels from the StyleSynth dataset. In the main text, shorter style names are used for presentation clarity.}
\label{tab:style-sense-activations}
\end{table*}

\begin{table}[h]
    \centering
    \resizebox{0.48\textwidth}{!}{
    \begin{tabular}{lccc}
    \toprule
    \textsc{Dataset} & \textsc{Split} & \textsc{Median Tokens} & \textsc{>128 Tokens} \\
    \midrule
    \multirow{2}{*}{StyleSynth} 
        & Train & 19 & 0.00\% \\
        & Test  & 10 & 0.00\% \\
    \midrule
    \multirow{2}{*}{Wegmann} 
        & Train & 24 & 6.78\% \\
        & Test  & 24 & 6.59\% \\
    \bottomrule
    \end{tabular}
    }
    \caption{Token-statistics across datasets and splits.}
    \label{tab:token_stats}
\end{table}

\subsection{Broader Applications}
\label{sec:appendix-broader-applications}
iBERT enables embedding-level interpretability and control, unlocking several downstream capabilities beyond benchmarking.

\paragraph{Style-Conditioned Retrieval.}
Rather than relying on keyword overlap or dense similarity, queries can be modified in latent space to match stylistic attributes (e.g., formality). This enables retrieval that is both semantically relevant and stylistically aligned—particularly useful in assistant and RAG pipelines where tone coherence matters.

\paragraph{Classifier Debiasing via Sense Ablation.}
Masking specific sense dimensions (e.g., sarcasm, punctuation cues) neutralizes stylistic noise before classification. This supports robust models that focus on core semantics—offering interpretability by construction.

\paragraph{Latent Data Augmentation.}
Controlled sense edits allow creation of style-shifted variants without decoding. These augmentations improve robustness and style-invariance for downstream classifiers.

\paragraph{Style-Aware Personalization.}
Sense-level edits adapt outputs to user preferences (e.g., concise vs. elaborate tone) without retraining—enabling personalization in retrieval and generation pipelines.

\paragraph{RAG and Retrieval Robustness.}
iBERT supports query normalization, document style filtering, and attribution for retrieval decisions—addressing failure modes like tone mismatch or prompt drift in RAG systems.

\paragraph{Forensic and Monitoring Applications.}
iBERT enables tracking tone shifts, authorial style evolution, or discourse change over time—useful for moderation, attribution, or sociolinguistic analysis.

\paragraph{Controller for Generation.}
Sense vectors from iBERT can guide decoder-based models (e.g., T5, GPT) as structured, interpretable control signals—supporting style-conditioned generation without prompt engineering.

\paragraph{Representation Auditing.}
Because embeddings decompose into interpretable senses, iBERT supports bias audits, attribution analysis, and fairness enforcement via projection or adversarial masking.

\paragraph{Zero-Decode Semantic Transfer.}
Sense-level edits enable style transfer, paraphrasing, or domain adaptation directly in latent space—without requiring any generation.

\subsection{Effect of Aggregation Sharpness}
\label{sec:res:tau}

\noindent
As described in Section \ref{sec:method}, $\tau$ controls how sharply the model aggregates across its $k{=}8$ sense vectors—ranging from uniform averaging ($\tau{\to}\infty$) to top-sense-only pooling ($\tau{=}0$). Intermediate values apply softmax pooling over sense magnitudes.

Sharper but soft aggregation version of v3 created by \(\tau{=}10\), v3-10 yields the best STEL and SoC performance, as reported in the full results (Table~\ref{tab:synthstel-45}). Appendix Table~\ref{tab:tau} highlights a representative subset across $\tau$ values, to draw attention to how aggregation sharpness affects performance and stability. This trend suggests that axis-selective pooling supports better alignment with stylistic dimensions. In addition to strong performance, \mbox{v3-10} also exhibits consistently lower variance across runs (see Table~\ref{tab:synthstel-45}), indicating more stable convergence during training. In contrast, hard top-sense selection degrades SoC, suggesting that some degree of sense blending is beneficial for generalization. These trends mirror regularization analogies: peaked pooling acts like L1, while smoother blending behaves L2-like.

\begin{table*}
\centering
\setlength{\tabcolsep}{4.5pt}
\resizebox{0.6\textwidth}{!}{
\begin{tabular}{@{}llcc|cc|cc@{}}
\toprule
& \multirow{2.5}{*}{\textsc{Variant}}
& \multicolumn{2}{c|}{\textsc{Frozen Phase}}
& \multicolumn{2}{c|}{\textsc{Unfrozen Phase}}
& \multicolumn{2}{c}{\textsc{Total}} \\
\cmidrule(lr){3-4}\cmidrule(lr){5-6}\cmidrule(lr){7-8}
& & Time & Loss / Epochs & Time & Loss / Epochs & Time & Epochs \\
\midrule
\multirow{2}{*}{\rotatebox{90}{\small 128}}
& iBERT & 23h   & 1.6912 / 57 & 3d 6h  & 1.3365 / 85 & 4d 5h  & 142 \\
& BERT  & 2h    & 1.3607 / 5  & 1d 2h  & 1.2228 / 31 & 1d 5h  & 36  \\
\addlinespace[4pt]
\midrule
\addlinespace[2pt]
\multirow{2}{*}{\rotatebox{90}{\small 512}}
& iBERT & 2d 9h & 1.4418 / 46 & 1d 7h  & 1.2941 / 16 & 3d 16h & 62  \\
& BERT  & 14h   & 1.1876 / 6  & 3d 8h  & 1.0940 / 46 & 3d 22h & 52 \\
\bottomrule
\end{tabular}
}
\caption{Training runtimes, epochs, and final losses for iBERT-MLM and BERT (MLM) under 128- and 512-token limits. Epoch counts exclude early-stopping patience rounds.}
\label{tab:train-times}
\end{table*}

\subsection{Effect of Pooling Sharpness \texorpdfstring{$\tau$}{tau}}
\label{app:tau}

To study the impact of pooling sharpness, we evaluated iBERT-Sentence (iSBERT) under four values of $\tau \in \{0, 1, 10, \infty\}$ referred to as iBERT-v1, iBERT-v2, iBERT-v3-1 and iBERT-v3-10.

We find that $\tau{=}0$ (v2, top-sense pooling) yields the highest sense-level sparsity, with sharper sense specialization and more effective style edits (\S\ref{sec:edit}). $\tau{=}1$ (v3, soft pooling) balances performance and interpretability. $\tau{=}10$ leans towards uniform weighting (v1) across senses.  $\tau{\to}\infty$ (v1, mean pooling) exhibits minimal sense-selectivity and lower editability. This confirms that sharper pooling enhances sense-disentanglement, critical for controllable edits and style probing.

\subsection{Training and Implementation Details}
\label{app:impl}

All iBERT models use \(k{=}8\) senses, 768-dim hidden states, and FlashAttention-2 \citep{dao2023flashattention2} for fast attention computation. 
AdamW optimizer with learning rate of ($2\!\times\!10^{-5}$), a batch size of 99 is used. The models have an early stopping with 3-epoch patience over validation loss and Flash-Attention-2 is used for faster compute. We perform both masked language modeling and sentence embedding training at input lengths of 128 and 512 tokens. MLM uses 5\% of FineWeb (750B tokens), and contrastive fine-tuning is performed using anchor-positive-negative triplets from StyleSynth / StyleDistance (SD) and Wegmann (WG) datasets.

\paragraph{Training.}
All models are initialized with encoder weights from ModernBERT \citep{warner-etal-2025-smarter}, replacing into either BERT or our proposed iBERT architecture. Due to the larger Context Sense Block in iBERT, we freeze the encoder weights in an initial warmup stage (for both baseline and iBERT to maintain parity) and allow the heads to converge. Once stabilized and converged, we unfreeze the full model and resume end-to-end training until convergence. This process is used in both MLM and contrastive phases. The details of the MLM stage are listed in \autoref{tab:train-times}.

Contrastive training is supervised with either InfoNCE or triplet loss, depending on the dataset. Triplets are grouped as $(\text{anchor}, \text{pos}, \text{neg})$ and processed using sentence encoders that implement SBERT-style wrappers over both BERT and iBERT backbones. Our training code supports sense ablation via \texttt{--sense\_gain}, and different SBERT pooling strategies (\texttt{v1}, \texttt{v2}, \texttt{v3-1}, \texttt{v3-10}).

Training is run on 4$\times$NVIDIA A100 (80GB) GPUs using mixed-precision (\texttt{torch.bfloat16}). We use AdamW with a learning rate of \(2 \times 10^{-5}\), batch size 99, and early stopping with 3-epoch patience on validation loss. All training stages are fully configurable via CLI or YAML using a shared interface, and can resume from checkpoints.

\paragraph{Software.}
Our implementation stack includes PyTorch (2.7.0), Transformers (v4.51.3), Sentence-Transformers (5.0.0), FlashAttention-2, Datasets (3.6.0), and Accelerate (1.7.0). Tokenization is handled via HuggingFace Tokenizers (0.21.1), and logging is supported through Terminal, JSONL, TensorBoard, and \text{Weights-and-Biases} (W\&B).

\paragraph{Data Statistics}
For \textsc{FineWeb}, we split the 5\% slice into a 90:10 train:dev split, without needing a test set, given it is used in an MLM. For the sentence-embedding training stages, we reuse the predefined train/test split provided by the dataset authors, and use 90:10 split of original train dataset for training and validation correspondingly.

\paragraph{Datasets, Licensing, and Safety}
Our study is based on publicly available datasets, which have been used in prior research on author attribution and stylistic analysis. These datasets originate from online platforms and may inherently contain offensive or personally identifiable content. However, we do not apply any additional filtering beyond what has been done in previous studies. To ensure ethical usage, we follow the guidelines set by the original sources that published these datasets and acknowledge any potential biases or content-related concerns that may arise.

Additionally, we discuss the licensing terms associated with the datasets used in our experiments. The datasets are provided under open-access licenses, allowing their use for academic research. We ensure compliance with these licenses and properly attribute the sources in our work. Detailed information on dataset licensing is included in the final version of the paper.

\subsubsection{Sentence Embedding Training and Replications}

For sentence-level contrastive training, we train a total of five distinct sentence encoder models (per data configuration): a baseline SBERT built on top of BERT, and four sentence encoders built on iBERT-MLM checkpoints using different pooling frameworks. These correspond to mean-pooling (v1, $\tau{=}\infty$), top-sense (v2, $\tau{=}0$), softmax-blend (v3-1, $\tau{=}1$), and peaked-blend (v3-10, $\tau{=}10$). Each is treated as a separate model, not as a parameter sweep. All models are trained at two input lengths (128 and 512) and evaluated on StyleSynth and Wegmann datasets.

To account for training variability, we perform 5 independent training replications for each sentence encoder: (S)BERT and all iBERT variants, at both token lengths (128 and 512). Reported results reflect mean and standard deviation across the runs.
\begin{align*}
\text{Training runs}
&= 5       && \left\{ \begin{array}{@{}l@{}}
  \text{model types:} \\
  \text{(S)BERT, iBERT-v1, v2} \\
  \text{iBERT-v3-1, v3-10}
  \end{array} \right. \\
&\times 3  && \left\{ \begin{array}{@{}l@{}}
  \text{datasets:} \\
  \text{SD, WG+SD, WG}
  \end{array} \right. \\
&\times 2  && \left\{ \begin{array}{@{}l@{}}
  \text{token lengths:} \\
  \text{128, 512}
  \end{array} \right. \\
&\times 5  && \left\{ \begin{array}{@{}l@{}}
  \text{replications per} \\
  \text{configuration}
  \end{array} \right. \\
&= \textbf{150}
\end{align*}

\subsection{Style Contrast Visualization}
\label{app:style-tsne}

We visualize the separation of contrastive style pairs (e.g., number substitution, personal pronoun frequency) using t-SNE projections. For each example, we compare (S)BERT embeddings (left) and iBERT-v2 embeddings (right). In both cases, sentence embeddings were obtained after training on StyleSynth.

Figure~\ref{fig:appendix-tsne} shows that iBERT exhibits stronger axis-wise separation between stylistic variants, with clearer clusters and lower embedding overlap, consistent with our claims about modularity and disentanglement.

\subsection{Global t-SNE for Embedding Clustering}
\label{app:global-tsne}
\begin{figure*}[htbp]
  \centering

  \begin{subfigure}{0.8\linewidth}
    \includegraphics[width=\linewidth,height=0.4\textheight,keepaspectratio]{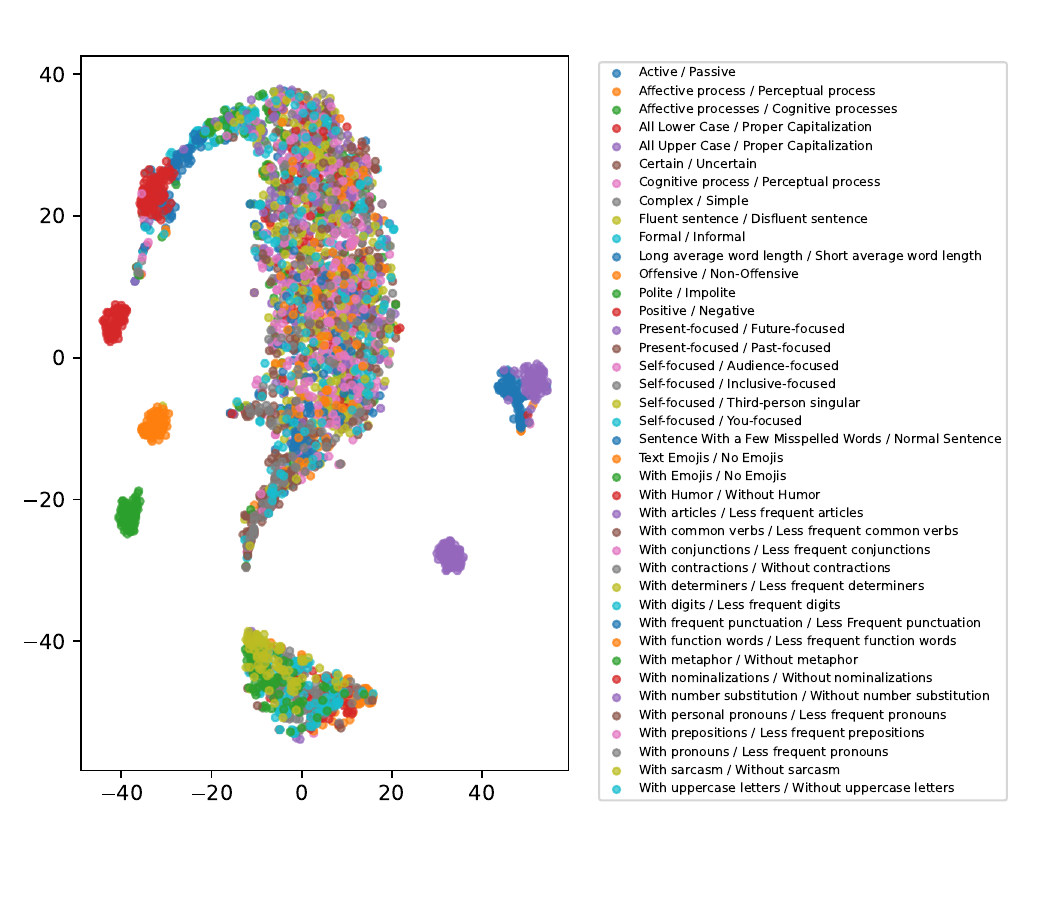}
    \caption{BERT (128 tokens)}
    \label{fig:tsne_vanilla}
  \end{subfigure}

  \vspace{1em}

  \begin{subfigure}{0.8\linewidth}
    \includegraphics[width=\linewidth,height=0.4\textheight,keepaspectratio]{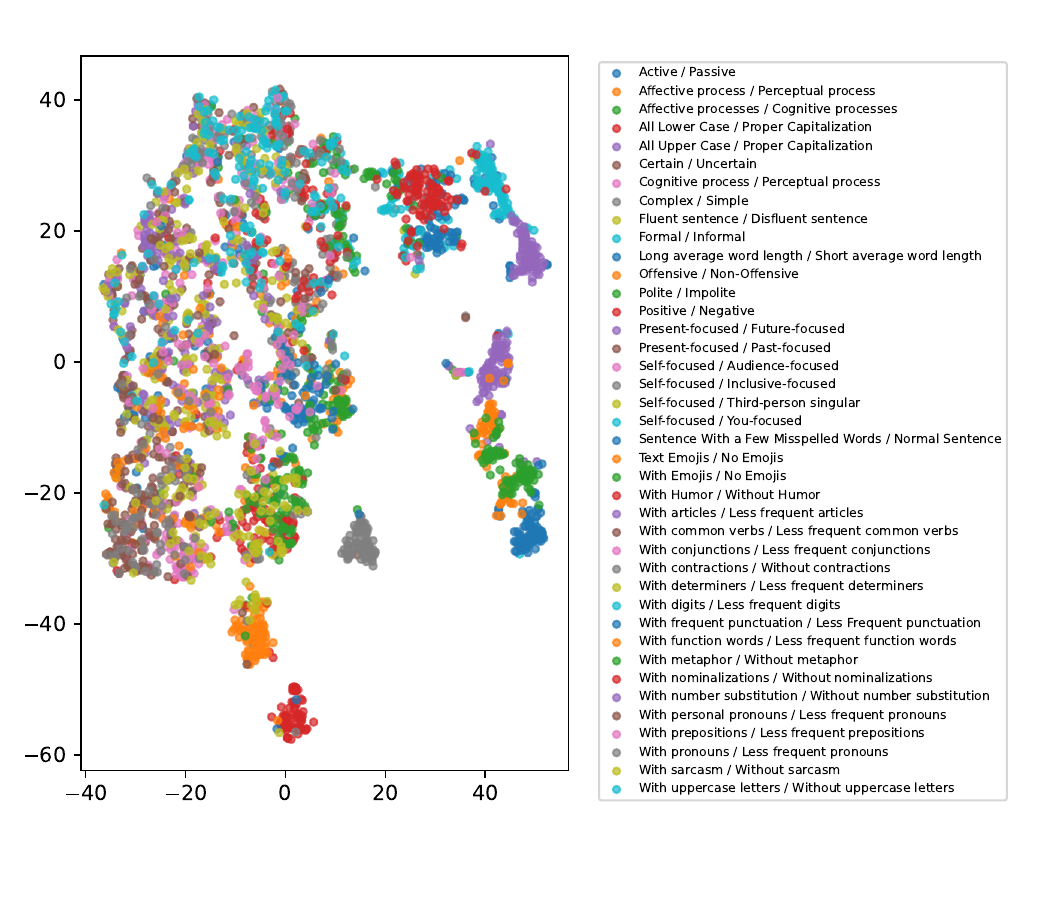}
    \caption{iBERT-v3-10 (128 tokens)}
    \label{fig:tsne_v3}
  \end{subfigure}

  \caption{t-SNE projections of sentence embeddings for all 40 style contrast pairs. Each point represents a sentence, color-coded by its style label. BERT shows more dispersed and overlapping clusters, while iBERT-v3-10 shows tighter, aligned groupings — indicating improved style modularity and disentanglement. Both these language models were first trained on \textsc{FineWeb} and then trained on StyleSynth triplets.}
  \label{fig:tsne_vanilla_v3}
\end{figure*}

We visualize sentence embeddings across all 40 style contrasts using t-SNE, comparing BERT and iBERT-v3 in Figure~\ref{fig:tsne_vanilla_v3}. The iBERT model exhibits sharper geometric structure, with more linear and segregated style clusters. This supports our hypothesis that modular pooling (v3) induces interpretable and axis-aligned representations, in contrast to the entangled embedding space of BERT.

\subsection{Model Schematic and Embeddings}
\vspace{0.5em}
We conclude the appendix with Figures~\ref{fig:iBERT-detailed} and~\ref{fig:tsne_vanilla_v3}, which illustrate the full model architecture and the global embedding structure.

\end{document}